\documentclass[sigconf]{acmart}
\AtBeginDocument{%
  \providecommand\BibTeX{{%
    \normalfont B\kern-0.5em{\scshape i\kern-0.25em b}\kern-0.8em\TeX}}}

\acmConference[ACM MM'22]{ACM Multimedia conference}{October 2022}{Lisbon Portugal}
\acmYear{2022}
\copyrightyear{2022}

\copyrightyear{2022} 
\acmYear{2022} 
\setcopyright{acmlicensed}\acmConference[MM '22]{Proceedings of the 30th ACM International Conference on Multimedia}{October 10--14, 2022}{Lisboa, Portugal}
\acmBooktitle{Proceedings of the 30th ACM International Conference on Multimedia (MM '22), October 10--14, 2022, Lisboa, Portugal}
\acmPrice{15.00}
\acmDOI{10.1145/3503161.3547846}
\acmISBN{978-1-4503-9203-7/22/10}

\newcommand{\etal}{\textit{et al.~}}
\usepackage{multirow}

\usepackage{algorithm}
\usepackage{algorithmicx}
\usepackage{algpseudocode}

\begin{document}

\title{Dynamic Graph Reasoning for Multi-person 3D Pose Estimation}

\author{Zhongwei Qiu}
\authornote{Equal Contribution}
\affiliation{%
  \institution{University of Science and Technology Beijing}
  \country{}}
\email{qiuzhongwei@xs.ustb.edu.cn}

\author{Qiansheng Yang}
\authornotemark[1]
\affiliation{%
  \institution{Baidu}
  \country{}}
\email{yangqiansheng@baidu.com}

\author{Jian Wang}
\affiliation{%
  \institution{Baidu}
  \country{}}
\email{wangjian33@baidu.com}

\author{Dongmei Fu}
\affiliation{%
\institution{University of Science and Technology Beijing}
  \country{}}
\email{fdm_ustb@ustb.edu.cn}
\begin{abstract}

Multi-person 3D pose estimation is a challenging task because of occlusion and depth ambiguity, especially in the cases of crowd scenes. To solve these problems, most existing methods explore modeling body context cues by enhancing feature representation with graph neural networks or adding structural constraints. However, these methods are not robust for their single-root formulation that decoding 3D poses from a root node with a pre-defined graph. In this paper, we propose GR-M3D, which models the \textbf{M}ulti-person \textbf{3D} pose estimation with dynamic \textbf{G}raph \textbf{R}easoning. The decoding graph in GR-M3D is predicted instead of pre-defined. In particular, It firstly generates several data maps and enhances them with a scale and depth aware refinement module (SDAR). Then multiple root keypoints and dense decoding paths for each person are estimated from these data maps. Based on them, dynamic decoding graphs are built by assigning path weights to the decoding paths, while the path weights are inferred from those enhanced data maps. And this process is named dynamic graph reasoning (DGR). Finally, the 3D poses are decoded according to dynamic decoding graphs for each detected person. GR-M3D can adjust the structure of the decoding graph implicitly by adopting soft path weights according to input data, which makes the decoding graphs be adaptive to different input persons to the best extent and more capable of handling occlusion and depth ambiguity than previous methods. We empirically show that the proposed bottom-up approach even outperforms top-down methods and achieves state-of-the-art results on three 3D pose datasets.
\end{abstract}




\maketitle

\section{Introduction}
\label{sec:intro}

\begin{figure}[t]
\centering
\includegraphics[width=0.99\columnwidth]{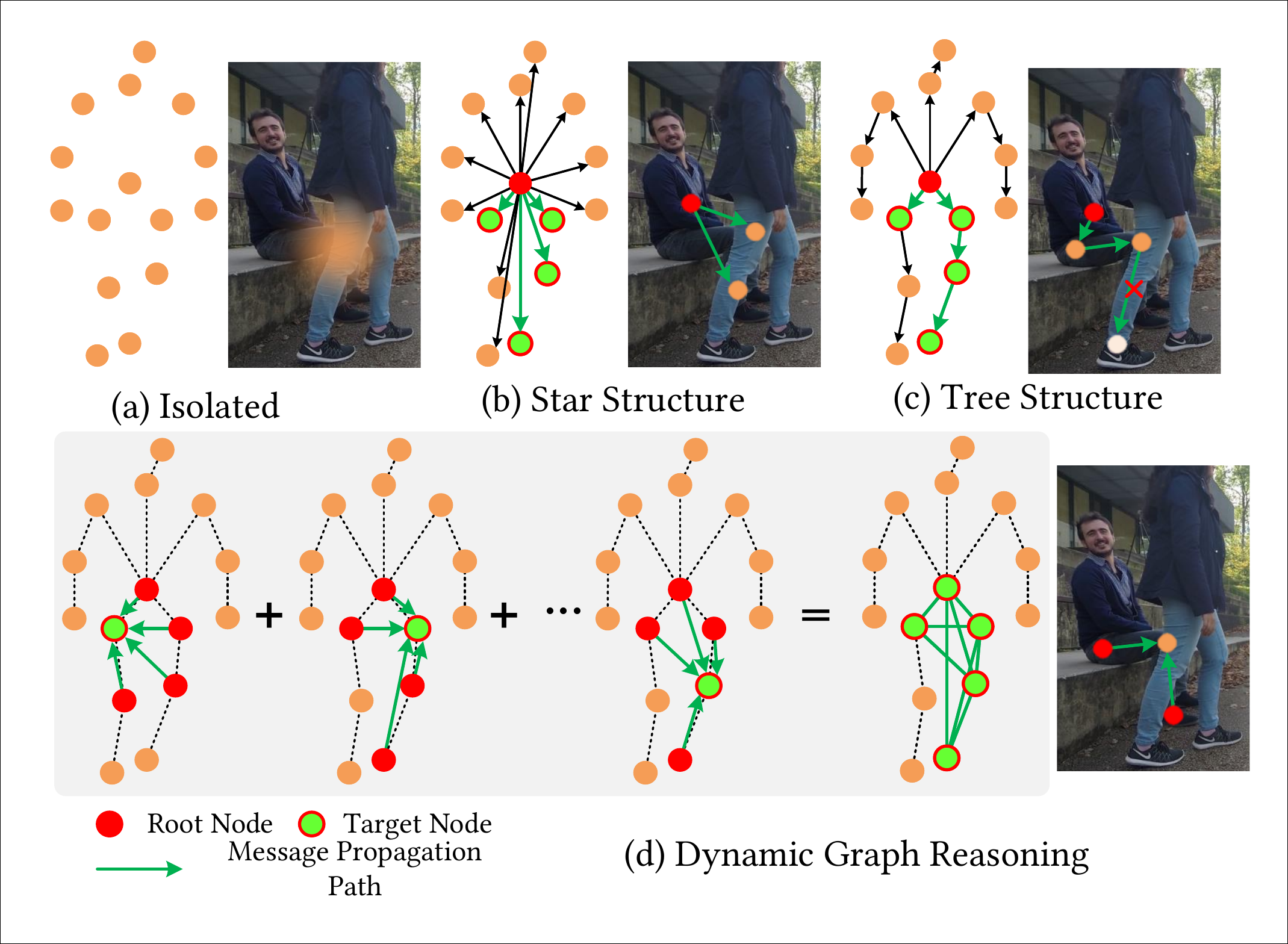} 
\caption{(a) Heatmap-based approaches~\cite{sun2018integral,moon2019camera} decode human poses by locating the point with maximum confidence on 3D heatmaps. Each node is isolated when decoding.
(b) Star graph approaches~\cite{nie2019single,zhou2019objects} start from single-root node, broadcasting information to other nodes by star graph, which suffers from long-distance information transmission problem.
(c) Tree graph approach~\cite{nie2019single} starts from single-root node, hierarchically transferring the message from the parent node to the child node, which suffers from the occlusion problem and cumulative error. 
(d) Our dynamic graph reasoning (DGR) extracts dense decoding paths from multiple root keypoints, further reasons the best dynamic decoding graphs for each person in the image from the predicted dense decoding paths. The final 3D poses are decoded with the guidance of reasoned dynamic decoding graphs.
}  
\label{fig1}
\end{figure}

The goal of multi-person 3D human pose estimation is to estimate the 3D coordinates of human joints from multiple human bodies in a monocular RGB image, which is a challenging and fundamental task. Recently, 3D human pose estimation has drawn a lot of attention because of its broad applications, such as human behavior understanding~\cite{yan2018spatial,liu2020disentangling}, human-object interaction detection~\cite{li2020detailed}, and athletic training assistance~\cite{wang2019ai}. Although remarkable progresses~\cite{martinez2017simple,sun2018integral,rogez2019lcr,moon2019camera,wang2020hmor} have been achieved in 3D pose estimation, the challenges of depth ambiguity and occlusions remain.

Existing multi-person 3D pose estimation methods mainly include top-down, bottom-up, and single-stage strategies. 
Top-down strategy~\cite{rogez2019lcr,rogez2019lcr++,moon2019camera,wang2020hmor} firstly predicts the human bounding box and the absolute depth of root points in each person and then conducts single-person 3D pose estimation in the bounding box. 
Bottom-up strategy~\cite{zanfir2018deep,mehta2018single,mehta2019xnect,zhen2020smap} firstly estimates the 3D coordinates for each human joint in an image and then assigns them to different human instances. Single-stage strategy~\cite{zhou2019objects,nie2019single} predicts keypoints offsets for each place in the image to generate 3D poses.
The top-down methods are more accurate but more costly since the human detection and the repeated stages of extracting features for each person.
The bottom-up and single-stage approaches are more efficient but uncompetitive in accuracy. The occlusions, non-uniform scales, and variable depths of each person are more difficult to handle when the input is a whole image, which are our interests.

For 3D human pose estimation, the common way to decode 3D pose is based on heatmaps ~\cite{sun2018integral,moon2019camera,wang2020hmor,zhen2020smap,ma2021context}, which decodes 3D coordinates from 3D heatmaps by an isolated structure as Figure \ref{fig1} (a).
To mitigate occlusion problem, \cite{zhen2020smap,ma2021context} model context information in coordinate level and feature level, respectively.
Recent works~\cite{zhou2019objects,nie2019single} bridge these two levels to some extent by estimating the 3D keypoints offsets, directly decoding keypoints coordinates from root point with 3D offsets. 
For example, based on single root point, CenterNet~\cite{zhou2019objects} and SPM~\cite{nie2019single} estimate other 3D keypoints by star structure propagation as Figure \ref{fig1} (b), which suffers the long-distance transmission problem. 
SPM further improves the information propagation path by tree structure decoding, as shown in Figure \ref{fig1} (c). 
The tree structure brings accumulation error since the hierarchical decoding scheme, which is serious when the middle nodes are occluded.

The single-root decoding strategy, such as star and tree structure decoding, is unreliable since the depth ambiguity.
Firstly, it is hard to calculate the localization of a new joint when its root joint is occluded. 
An incorrect starting point will result in a wrong whole 3D pose prediction. Secondly, the predicted depth from single root points is not robust since lacking the global context of the human body. Thirdly, the decoding graph is fixed for each person, which may fail to build the message propagation path when the middle nodes are occluded. 

Based on the above analysis, we propose a dynamic graph reasoning method (DGR). As shown in Figure \ref{fig1} (d), each joint in a person serves as a root keypoint to generate dense decoding paths. 
DGR reasons the dynamic decoding graph for each person from these dense decoding paths.
Each message propagation path in the dynamic decoding graph is calculated by combining starting point, target point, and message path score, which is adaptively adjusted according to the occlusions in the input image. 
This mechanism enables the best decoding paths for each target joint.
Although some works~\cite{zhao2019semantic,zeng2020srnet,ma2021context} use graph convolutional network or local human structure to learn better structural features, the learned body graph is also fixed. 
In the inference phase, this learned structural graph can't be adjusted adaptively with the occlusion of the input image. But our DGR can self-adaptively adjust the decoding graph due to the multi-root decoding and dynamic graph reasoning mechanism.

To better predict the depth and build a reliable message propagation graph, we further propose a scale and depth aware refinement (SDAR) module. Inspired by the intuition of near big far small, which is a basic perspective principle, SADR concatenates multiple initial scale features and depth features and then generates refined scale and depth features. The refined scale-aware feature and depth-aware feature are beneficial to generating good root keypoints and building robust pose graphs.

Our main contributions can be summarized as follows: 
\begin{itemize}
    \item We argue that decoding 3D poses from a single root point is unreliable and propose a novel dynamic graph reasoning (DGR) method to decode multi-person 3D poses.
    \item We propose SADR to generate better root keypoints and reliable decoding graph representations by aggregating global depth and scale information.
    \item The proposed bottom-up approach outperforms even top-down methods and achieves new state-of-the-art results on three widely-used benchmarks: Human3.6M, MuPoTS-3D, and CMU Panoptic datasets.
\end{itemize}

\section{Related Work}

\subsection{2D Human Pose Estimation}
Mainstream multi-person 2D pose estimation approaches include top-down and bottom-up methods. Top-down approaches~\cite{chen2018cascaded,xiao2018simple,sun2019deep} firstly conduct human detection. Then, they crop images and perform single-person human pose estimation for each human instance. 
Xiao~\etal~\cite{xiao2018simple} propose a simple baseline for 2D pose estimation, which uses ResNet as the backbone and follows several up-sample layers to generate heatmaps. Sun~\etal~\cite{sun2019deep} propose HRNet to generate high-resolution representation. Bottom-up approaches~\cite{cao2017realtime,kocabas2018multiposenet,kreiss2019pifpaf} estimate the keypoints for all human instances in an image and then group them into multiple instances. Cao~\etal~\cite{cao2017realtime} propose part affinity fields to group keypoints. Kocabas~\etal~\cite{kocabas2018multiposenet} propose Multiposenet, a framework to finish detection, pose estimation, and grouping at the same time. The accuracy of bottom-up approaches is lower than top-down approaches since the different scales and low resolution of persons. 

Some approaches~\cite{nie2019single,zhou2019objects,tian2019directpose,wei2020point} discard the method of locating from heatmaps, but learn 2D offset to decode pose. Nie~\etal~\cite{nie2019single} propose SPM to predict root points of the human body and offsets of each keypoints. The coordinates of keypoints can be obtained from the root points and offsets. DirectPose~\cite{tian2019directpose} and Point-set anchors~\cite{wei2020point} use deformable convolution to align the features of pose. These offset-based methods provide insight for 3D pose estimation.

\subsection{3D Human Pose Estimation}
For single person cases, the main-stream methods follow the architecture of top-down 2D pose estimation methods. They change the 2D pose regression heads to 3D heads, including \textit{inferring} and \textit{lifting} methods. The \textit{inferring} methods~\cite{pavlakos2017coarse,mehta2017monocular,sun2017compositional,kanazawa2018end,sun2018integral} directly regress 3D pose from learned 3D heatmaps. For example, Sun~\etal~\cite{sun2018integral} conduct integral operation on 3D heatmaps to obtain 3D pose coordinates. The \textit{lifting} methods~\cite{martinez2017simple,moreno20173d,yang20183d,zeng2020srnet} firstly estimate 2D pose by 2D pose estimator, and then lift 2D pose to 3D pose by a sample neural network, such as SRNet~\cite{zeng2020srnet}.

Recently, some works~\cite{rogez2019lcr,rogez2019lcr++,zanfir2018deep,moon2019camera,wang2020hmor} study multi-person 3D pose estimation.
Rogez~\etal~\cite{rogez2019lcr,rogez2019lcr++} locate the human bounding box and generate a set of anchor poses for each human. The anchor poses are refined to the final pose by a regression module. 
Moon~\etal~\cite{moon2019camera} propose a top-down pipeline as multi-person 2D pose estimation. They locate the human bounding box and the depth of root keypoints in each bounding box and then conduct single-person 3D pose estimation in the bounding box.
Zanfir~\etal~\cite{zanfir2018deep} propose MubyNet, which estimates keypoints and limb core at the same time and then integrates limb score to group keypoints into different persons. 
Wang~\etal~\cite{wang2020hmor} propose hierarchical multi-person ordinal relations as an additional loss for depth learning. 
Cheng~\etal~\cite{cheng2021monocular} integrate top-down and bottom-up networks to relieve the problems of occlusion and close interactions.

\subsection{Graph Reasoning in Pose Estimation}
Since human joints can be naturally seemed as a graph structure, many works~\cite{ci2019optimizing,qiu2019learning,nie2019single,cai2019exploiting,zhao2019semantic,qiu2020dgcn,ma2021context} try to utilize this information. 
Zhao~\etal~\cite{zhao2019semantic} take the human joints as the nodes of graph and then build a semantic graph convolutional network to learn joints relation. More works try to learn a better graph representation for human pose, such as spatial-temporal graph convolutional network~\cite{cai2019exploiting}, dynamic graph convolutional network~\cite{qiu2020dgcn}, context pose~\cite{ma2021context} and so on. 
Other works~\cite{ci2019optimizing,zeng2020srnet} study the local or global structure of the human body. However, these methods just learn graph representation for the human pose, the learned human graph is fixed in the inference phase. In this paper, different from previous works, we propose a graph reasoning approach, which can self-adaptively adjust the graph in the inference phase.

\begin{figure*}[t]
\centering
\includegraphics[width=0.99\textwidth]{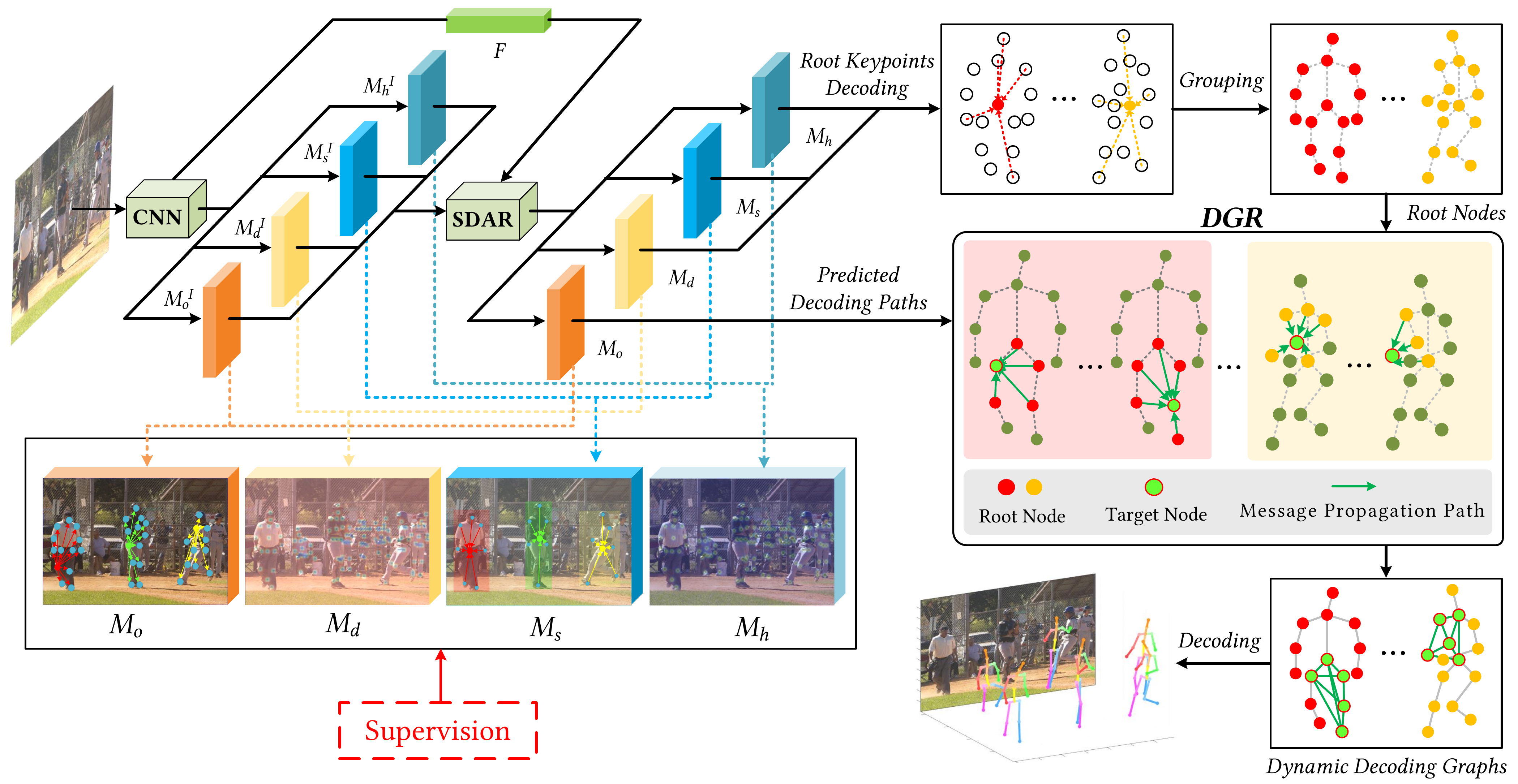} 
\vspace{-0.3cm}
\caption{The framework of our approach includes: 1) Given an image, backbone network processes deep feature ($F$) and four maps ($M^I_h$, $M^I_s$, $M^I_d$, $M^I_o$). 2) Scale and Depth Aware Refinement (SDAR) module aggregates scale and depth information to enhance feature $F$ and refine the four maps with the scale and depth context. 3) Multiple root keypoints are decoded from $M_h, M_s, M_d$ by Multi-person Root Keypoints Decoding (MRKD). 4) Based on the multi-root keypoints, we conduct dynamic graph reasoning on predicted dense paths to generate dynamic decoding graphs for each person, to further decode final 3D poses.}
\label{framework}
\end{figure*}

\section{Approach}
Let $p$ denotes the coordinate of a target joint, then the offset-based methods~\cite{zhou2019objects,nie2019single} can be formulated as:
\begin{equation}
\label{Equ_Single-Root}
    p = p^r + \Delta{p}
\end{equation}
where $p^r$ is the coordinate of a root joint, such as body center or the parent joint of $p$ based on the human skeleton. $\Delta{p}$ is a learned offset from the root joint to the target joint.

However, predicting 3D pose from single root is not robust since depth ambiguities and occlusions. Thus, we propose a new graph-based method, which takes full advantage of the context, scale, and depth information of multiple roots, and makes the 3D pose prediction more robust for handing depth ambiguity and occlusion.

\subsection{Framework} \label{Framework}

The overview of our dynamic graph reasoning is illustrated in Figure \ref{framework}. First of all, four maps and deep feature map $\textbf{F}$ are extracted from the backbone network. 
These maps are denoted as $\textbf{M}_h^I$, $\textbf{M}_d^I$, $\textbf{M}_s^I$ and $\textbf{M}_o^I$ of shapes $(K+1, H, W)$, $(K, H, W)$, $(2, H, W)$ and $(3K, H, W)$ respectively.
$K$ is the number of joint categories, $H$ and $W$ are the height and width of these maps. $\textbf{M}_h^I$ is the heat map for $K$ joints and one body center (the midpoint of left hip and right hip). 
$\textbf{M}_d^I$ is the depth map for $K$ joints. 
$\textbf{M}_s^I$ is a map where each pixel preserves the 2D offset from itself to the corresponding body center. And it can be regarded as scale map since the 2D offset indicates the scale of a person. 
$\textbf{M}_o^I$ is a 3D offset map, where each pixel records the 3D offsets from its 3D location to the corresponding $K$ joints.

Based on those maps, our method outputs multi-person 3D pose by carrying out the following three parts successively, which are Scale and Depth Aware Refinement (SDAR), Multi-person Root Keypoints Decoding (MRKD), and Dynamic Graph Reasoning (DGR). Firstly, SDAR refines the four maps mentioned above by integrating scale and depth information, and the refined maps are denoted as $\textbf{M}_h$, $\textbf{M}_d$, $\textbf{M}_s$, $\textbf{M}_o$. 
Secondly, MRKD decodes multi-person root keypoints from those refined maps. 
$\textbf{M}_h$, $\textbf{M}_d$, $\textbf{M}_s$ are used to decode 3D root keypoints and $\textbf{M}_o$ is used to decode dense decoding paths.
Finally, based on the root keypoints and predicted dense decoding paths, DGR reasons the dynamic decoding graphs for each person in the input image, to further decode robust multi-person 3D poses as equation \ref{Equ_Single-Root} with the guidance of dynamic decoding graphs. 
The details of the three parts and the training process are introduced in the following.

\subsection{Scale and Depth Aware Refinement} 
\label{SDAR}
\begin{figure}[t]
\centering
\includegraphics[width=0.99\columnwidth]{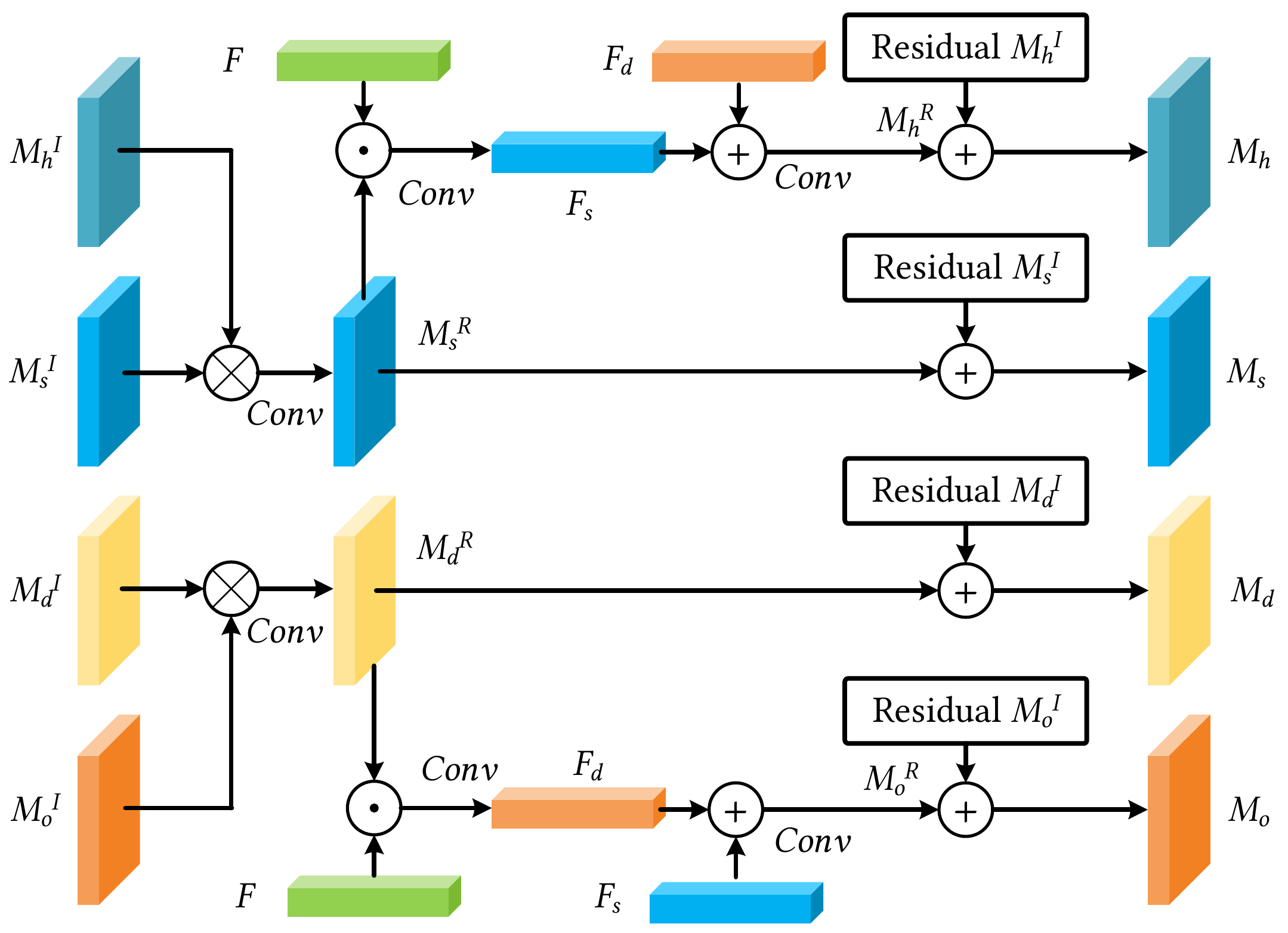} 
\caption{The architecture of scale and depth refinement (SDAR). SDAR aggregates scale and depth information to learn scale-aware and depth-aware maps, further to refine keypoints heatmaps and 3D offsets. $\otimes$ denotes concatenate operation, $\oplus$ denotes element-wise addition, $\odot$ denotes element-wise multiplication.
}  
\label{fig_sdar}
\end{figure}

As demonstrated in Figure \ref{fig_sdar}, SDAR firstly computes refined scale map and depth map by the following operations:
\begin{align}
    \textbf{M}_s^R &= Conv_s(Concat(\textbf{M}_h^I, \textbf{M}_s^I))    \label{Equ_ScaleMap}\\
    \textbf{M}_d^R &= Conv_d(Concat(\textbf{M}_o^I, \textbf{M}_d^I))    \label{Equ_DepthMap}
\end{align}
where $Conv_s$ and $Conv_d$ are sequential modules including convolution and normalization layers, and $Concat$ is an operator for concatenating tensors in channel dimension. Equation \ref{Equ_ScaleMap} merges heat map into scale map, which makes $\textbf{M}_s$ can provide both scale information and attention cue for downstream operations.
In Equation \ref{Equ_DepthMap}, depth map is refined to capture more global depth context from $M_o^I$.

Once got the refined scale map and depth map, we multiply them with the feature map $\textbf{F}$ respectively, and then sum up these products as $\textbf{F}_{sd}$. Based on this enhanced feature, the refined heat map and 3D offset map can be predicted by the following calculation:
\begin{equation}
    \begin{aligned}
        \textbf{M}_h^R, \textbf{M}_o^R &= Split(Conv_f(\textbf{F}_{sd}))\\
        \textbf{F}_{sd} &= \textbf{M}_s \odot \textbf{F} + \textbf{M}_d \odot \textbf{F}    \label{Equ_HeatMap_OffsetMap}
    \end{aligned}
\end{equation}
where $Conv_f$ is also a sequential convolution module, $Split$ is an operator for splitting tensor in channel dimension, and $\odot$ means element-wise multiplication.

The final outputs are:
\begin{equation}
    \textbf{M}_* = \textbf{M}_*^I + \textbf{M}_*^R
\end{equation}
where $*$ is wildcard character for $\{h, s, d, o\}$. 

In the following, $\textbf{M}_h$, $\textbf{M}_s$ and $\textbf{M}_d$ will serve for multi-person root keypoints decoding, while $\textbf{M}_h$ and $\textbf{M}_o$ will contribute for dynamic graph reasoning. After the refining by SDAR as Equation \ref{Equ_HeatMap_OffsetMap}, the final multi-person pose predicted by DGR could be more precise since the heatmap and 3D offset map had been enhanced by scale and depth references.

\subsection{Multi-person Root Keypoints Decoding} 
\label{MRKD}
Before carrying out graph reasoning, we decode multi-person root 3D keypoints out of the maps outputted from SDAR. This process is called multi-person root keypoints decoding (MRKD). 
At first, $N\times K$ independent keypoints and $N$ body centers can be detected from $\textbf{M}_h$ by extracting local maximums, where $N$ represents the number of persons and is obtained by finding $N$ high-confidence points according to a confidence threshold. 
Meanwhile, the depth of each keypoint is the value of $\textbf{M}_d$ at the corresponding 2D location. 
And then, we assign these keypoints to $N$ persons.
Let $[p_1,...,p_m,...,p_M]$ as the $M=N\times K$ predicted keypoints and $[c_1,...,c_n,...,c_N]$ as the $N$ predicted body centers from $\textbf{M}_h$, where $p_m$ and $c_n$ are both 2D coordinates.
Then each $p_m$ is corresponding to a unique $c_n$ by solving a distance matrix $E^{M\times N}$ with hungarian algorithm, while the definition of each element in $E$ is:
\begin{equation}
    \begin{aligned}
        E[m, n] &= \mathcal{E}(\widetilde{c}_m, c_n)\\
        \widetilde{c}_m &= \textbf{M}_s|_{p_m} + p_m
    \end{aligned}
\end{equation}
where $\mathcal{E}$ represents calculating euclidean distance, and $\widetilde{c}_m$ is a regressed 2D body center coordinate from $p_m$ depending on the semantic information at $p_m$, while $\textbf{M}_s|_{p_m}$ means the 2D offset from $p_m$ to this center, which can be assigned as the 2D vector at point $p_m$ on $\textbf{M}_s$.

\subsection{Dynamic Graph Reasoning} 
\label{DGR}
After obtaining root keypoints, the decoding graphs are inferred by conducting dynamic graph reasoning (DGR) for the final 3D pose estimation.
Supposing that a person can be regarded as a undirected acyclic graph, denoted as $\mathbf{G}(\mathbb{P}, \mathbb{E})$, where $\mathbb{P} = \{ p^i , i \in [1,K] \}$ is the set of joints for this person, $p^i$ is the 2D coordinate of the $i^{th}$ joint. Here $\mathbb{P}$ is initialized with the root joints detected from $M_h$ and $M_s$. While, $\mathbb{E} = \{e^{ij}, i\in [1,K], j \in [1,K]\}$ is the set of dense decoding paths and $e^{ij}$ means the decoding path from the $i^{th}$ joint to the $j^{th}$ joint. And it is valued by the 3D offset which is expressed as:
 
\begin{equation}
    e^{ij} = \textbf{M}_o|_{p^i}^j
\end{equation}
where $\textbf{M}_o|_{p^i}^j$ means the 3D offset to the $j^{th}$ target joint from $\textbf{M}_o$ at the position of root joint $p^i$.

The goal of DGR is to reason the best decoding paths $\hat{\mathbb{E}}$ from dense decoding paths set $\mathbb{E}$, to further construct decoding graph $\mathbf{G}(\mathbb{P}, \hat{\mathbb{E}})$ for each person in the input image. Intuitively, we can directly pick or drop candidate paths for generating $\hat{\mathbb{E}}$. However, this hard selecting manner may not be optimal. Here, we adopt a soft manner that assigning a weight on each candidate path for reasoning the decoding graph. And the weights for all paths are inferred from heat map $\textbf{M}_h$ and offset map $\textbf{M}_o$. While the weighted decoding paths can be expressed as
\begin{equation}
    \hat{\mathbb{E}} = \{(e^{ij}, \mathcal{W}(p^i, p^j)), i\in [1,K], j \in [1,K]\}
\end{equation}
where $\mathcal{W}(p^i, p^j)$ means the corresponding path weight for $e^{ij}$, which can be calculated as:
\begin{equation}
    \mathcal{W}(p^i, p^j) = \textbf{M}_h|_{p^i}^i\mathcal{R}(p^i, p^j)\textbf{M}_h|_{p^j}^j \label{Equ_Message}
\end{equation}
where $\textbf{M}_h|_{p^i}^i$ and $\textbf{M}_h|_{p^j}^j$ serve as confidence scores which are the heat values of the $i^{th}$ joint at point $p^i$ on $\textbf{M}_h$ and the $j^{th}$ joint at point $p^j$. While $\mathcal{R}(p^i, p^j)$ is a bone confidence formulated as:
\begin{equation}
    R({p}^{i},{p}^{j}) = \exp(-(\frac{||\textbf{M}_o|_{p^i}^j||_2}{||\textbf{M}_o|_{p^h}^{c}||_2}+\gamma(i,j)))
\end{equation}
where $h$ and $c$ mean the joint index of head-top and mid-hip, respectively. $\textbf{M}_o|_{p^h}^{c}$ is the 3D offset from the head-top to mid-hip joint. $\gamma(i,j) = ||\frac{\sigma (i,j)}{\sigma (h,c)}||_2$ is a priori ratio, and $\sigma (i,j)$ is the average bone length between the $i^{th}$ and the $j^{th}$ joint, counted from the training dataset, while $\sigma (h,c)$ is the average bone length between the $h^{th}$ joint and the $c^{th}$ joint. Thus, $R({p}^{i},{p}^{j})$ indicates the confidence of predicted edge in the human structural graph, which appears as a production of instance-level propagation confidence and statistical priori propagation confidence.

Given a pair of predicted root joint set $\mathbb{P}$ and decoding graph $\mathbf{G}(\mathbb{P}, \hat{\mathbb{E}})$, the final 3D pose of a certain person can be decoded. Concretely, we firstly extend $p^i$ in $\mathbb{P}$ as $p_{3d}^i$ by concatenating $p^i$ with its corresponding depth $d^i$ predicted from $M_d$. And the extended root joint set is denoted as $\mathbb{P}_{3d}$. Then we decode $\mathbf{G}(\mathbb{P}_{3d}, \hat{\mathbb{E}})$ and update the 3D coordinates of each joint as:
\begin{equation}
    \hat{p}_{3d}^j = \frac{\sum_i^K \mathcal{W}(p^i, p^j)(p_{3d}^i + e^{ij})}{\sum_i^K \mathcal{W}(p^i, p^j)}, \quad j = [1,K]
\end{equation}
where $\hat{p}_{3d}^j$ represents the estimated 3D coordinates of the $j^{th}$ target joint.

There are two reasons that the DGR can predict better 3D poses: 1) Compared with the single-root decoding mechanism, the depth value predicted by the multi-root decoding mechanism are more robust. 2) For each person, DGR builds a dynamic decoding graph, which can be adjusted by the path weights according to the deep feature of input image adaptively, even in the inference phase. As stated in Equation \ref{Equ_Message}, the graph path confidence is determined by start point confidence, edge confidence, and endpoint confidence, which perceives the occlusion of the inputs image. This makes the 3D pose decoding more robust to occlusion conditions.

\subsection{Loss Function} \label{Training} 
For training the network, we impose supervision both on the input and output maps of SDAR. The total loss is :
\begin{equation}
    \mathscr{L} = \mathscr{L}_{h} + \alpha \mathscr{L}_{s} + \beta \mathscr{L}_{d} + \mathscr{L}_{o},
\end{equation}
where $\alpha=0.1$ and $\beta=0.1$ are loss weights. and
\begin{equation}
    \begin{aligned}
        \mathscr{L}_h &= \mathcal{L}_{MSE}(\textbf{M}_h^I, \overline{\textbf{M}}_h) + \mathcal{L}_{MSE}(\textbf{M}_h, \overline{\textbf{M}}_h)\\
        \mathscr{L}_s &= \mathcal{L}_{L1}(\textbf{M}_s^I, \overline{\textbf{M}}_s) + \mathcal{L}_{L1}(\textbf{M}_s, \overline{\textbf{M}}_s)\\
        \mathscr{L}_d &= \mathcal{L}_{L1}(\delta(\textbf{M}_d^I), \overline{\textbf{M}}_d) + \mathcal{L}_{L1}(\delta(\textbf{M}_d), \overline{\textbf{M}}_d)\\
        \mathscr{L}_o &= \mathcal{L}_{L1}(\textbf{M}_o^I, \overline{\textbf{M}}_o) + \mathcal{L}_{L1}(\textbf{M}_o, \overline{\textbf{M}}_o)
    \end{aligned}
\end{equation}
where $\overline{\textbf{M}}_*$ represents ground truth map. $\mathcal{L}_{MSE}$ is standard MSE loss.
$\mathcal{L}_{L1}$ is L1 loss, and only pixels around body joints on $\overline{\textbf{M}}_s$, $\overline{\textbf{M}}_d$ and $\overline{\textbf{M}}_o$ are active for training. For training depth map, output transformation $\delta(x) = 1/sigmoid(x) - 1$ is applied on $\textbf{M}_d^I$ and $\textbf{M}_d$ before computing loss, following \cite{NIPS2014_7bccfde7}.

\section{Experiments}
\subsection{Datasets and evaluation metrics}
\textbf{MuCo-3DHP and MuPoTS-3D}
MuCo-3DHP is a large-scale indoor multi-person training dataset~\cite{mehta2017monocular}, including 400K frames for training. For testing, MuPoTS-3D, which consists of real images with various camera poses, are collected from indoor and outdoor scenes. There are 20 real-world scenes in the MuPoTS-3D dataset, which are labeled with ground-truth 3D poses for multiple person subjects, making it a convincing benchmark to test the generalization ability of the 3D pose model. 

The widely-used evaluation metric for multi-person 3D pose estimation is $3DPCK$. If the Euclidean distance between predicted and ground-truth is smaller than the threshold (150mm), the prediction is marked as a correct prediction. $PCK_{rel}$ measures relative pose accuracy after root alignment, and $PCK_{abs}$ measures absolute pose accuracy without root alignment. The area under the curve of $3DPCK$ over various thresholds is defined as $AUC$. To evaluate in the crowded scenes, we use \textit{Crowd Index} to split hard cases and easy cases. $0 \leq \textit{Crowd Index} < 1$ is introduced in \cite{li2019crowdpose}. The bigger \textit{Crowd Index} means the serious occlusion by human bodies.

\begin{table}[t]
\centering
\renewcommand\tabcolsep{2pt}
\caption{The ablation study of SDAR and DGR on MuPoTS-3D dataset. The star and tree decoding approaches are introduced in \cite{nie2019single}. The backbone of all results in this table is ResNet-34.}
\vspace{-0.3cm}
\resizebox{.99\columnwidth}{!}{
\begin{tabular}{lccccc}
\toprule
Model & Enhance Feature & Decoding Graph & 3DPCK& $\Delta$ \\
\midrule
Baseline & $\times $ & Star & 75.4 & -\\
Baseline & $\times $ & Tree & 76.7 & $\uparrow$1.7$\%$\\

Ours(GR-M3D) & $\times $ & DGR &77.9 & $\uparrow$3.3$\%$  \\
Ours(GR-M3D) & SDAR & DGR & \textbf{78.6} &$\uparrow$4.2$\%$  \\
\bottomrule
\end{tabular}}

\label{table_ablation_modules}
\end{table}

\begin{figure*}[t]
\centering
\includegraphics[width=0.95\textwidth]{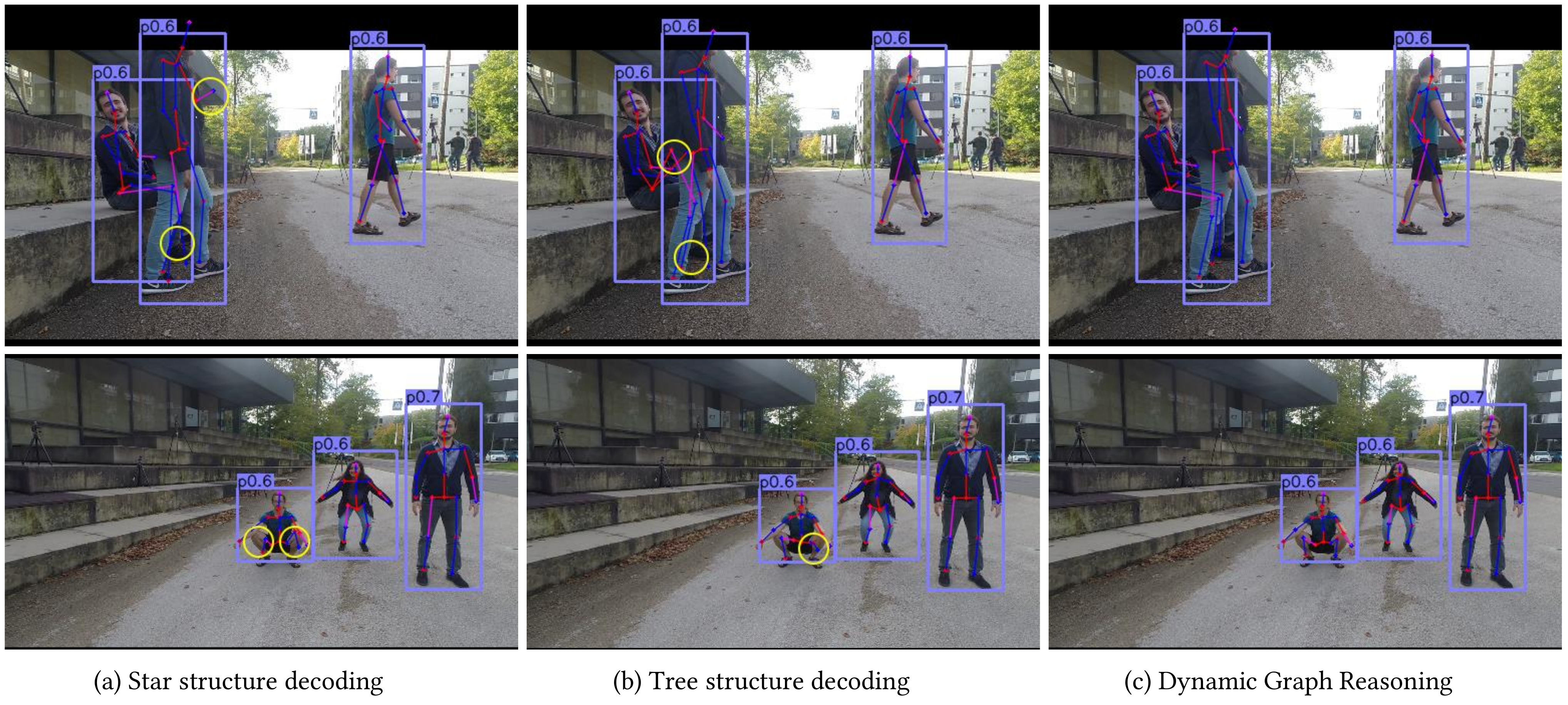} 
\vspace{-0.4cm}
\caption{The comparison of (a) star structure decoding, (b) tree structure decoding, and (c) our dynamic graph reasoning (DGR). DGR generates better results on these cases with occlusions or strange poses than star and tree structure decoding.}
\label{fig:case_ana}
\end{figure*}

\begin{table}[t]
\centering
\renewcommand\tabcolsep{2pt}

\caption{The ablation study of our approach (GR-M3D) based on different backbone on MuPoTS-3D. HG2 denotes Hourglass with 2 blocks. RN denotes ResNet. HRN32 denotes HRNet-32. Baseline is with star structure.}
\vspace{-0.3cm}
\resizebox{.99\columnwidth}{!}{
\begin{tabular}{lcccccc}
\toprule
Model & Backbone & Params(MB) & Time(s/img) & 3DPCK & $\Delta$\\
\midrule
Baseline & \multirow{2}{*}{RN34} &32.1 & 0.022 &75.4 & \multirow{2}{*}{$\uparrow$~4.2$\%$} \\  
GR-M3D &  & 34.4 & 0.026 & 78.6 & \\  
\midrule
Baseline & \multirow{2}{*}{RN50} & 40.6 & 0.029 &77.1 & \multirow{2}{*}{$\uparrow$~3.1$\%$} \\ 
GR-M3D &  & 42.9 & 0.034 & 79.5 & \\ 
\midrule
Baseline & \multirow{2}{*}{RN101} &59.6 &0.041  &79.1 & \multirow{2}{*}{$\uparrow$~2.8$\%$} \\  
GR-M3D &  & 61.9 & 0.045 & 81.3 & \\ 
\midrule
Baseline & \multirow{2}{*}{HRN32} & 40.1  & 0.061  & 80.6 & \multirow{2}{*}{$\uparrow$~2.9$\%$} \\ 
GR-M3D &  & 42.4 & 0.065 & 82.9 \\  
\midrule
Baseline & \multirow{2}{*}{HG2}  &192.1 & 0.151  & 82.2 & \multirow{2}{*}{$\uparrow$~2.9$\%$} \\
GR-M3D &   & 196.6 & 0.158 & \textbf{84.6} \\
\bottomrule
\end{tabular}}

\label{table_backbones}
\end{table}

\begin{table}[t]
\centering
\renewcommand\tabcolsep{3pt}
\caption{The ablation study of GR-M3D on MuPoTS-3D dataset with different crowd indexes. Backbone is ResNet-50 here. Baseline is with star structure.}
\vspace{-0.3cm}
\resizebox{.99\columnwidth}{!}{
\begin{tabular}{lccccc}
\toprule
Crowd Index & $>0.0$ & $>0.3$ & $>0.5$ & $>0.7$\\
\midrule
Baseline & 77.1 & 76.4 & 74.8 & 73.2\\
GR-M3D & \textbf{79.5}$\uparrow$2.7\% & \textbf{78.8}$\uparrow$3.1\% & \textbf{77.9}$\uparrow$4.1\% & \textbf{77.5}$\uparrow$5.9\% \\
\bottomrule
\end{tabular}}
\label{table_crowd_index}
\end{table}

\textbf{Human3.6M}
Human3.6M~\cite{ionescu2013human3} is the largest indoor benchmark for single-person 3D pose estimation, which consists of 3.6M video frames. Collectors use the motion capture system to obtain the ground-truth 3D poses.
MPJPE and PA-MPJPE are widely used to measure the accuracy of the 3D root-relative pose. They calculate the euclidean distance between predicted and ground-truth 3D joint coordinates after root joint alignment and further rigid alignment (i.e., Procrustes analysis~\cite{gower1975generalized}). 

\textbf{CMU Panoptic}
CMU Panoptic \cite{joo2017panoptic} is a large-scale multi-person 3D human pose estimation dataset, captured by multiple cameras. It's challenging to recover the multi-person 3D pose since heavy mutual occlusion. Following the settings of \cite{zanfir2018monocular,zhen2020smap}, 9600 images of two cameras (16, 30) from four activities (Haggling, Mafia, Ultimatum, Pizza) as the testing set, and 160k images from different videos as training set. For fair comparison with \cite{zanfir2018monocular,zhen2020smap}, MPJPE is the evaluation metric.

\subsection{Implement details}
GR-M3D is trained on 8 V100 GPUs with a batch size of 16/GPU and input resolution is $512\times 512$. Adam optimizer is adopted and the initial learning rate is 5e-4, which decreases 10× at 60 and 90 epochs. The total epoch number is 110. Random flip, random occlusion, rotation, and color jittering are used, and the range of rotation is $[-\pi, \pi]$. The confidence threshold for root ketpoints is 0.5. Unless otherwise specified, the backbone of GR-M3D is Hourglass network.
Following previous works~\cite{moon2019camera,fabbri2020compressed,zhen2020smap}, additional 2D images in MPII~\cite{andriluka20142d} and COCO~\cite{lin2014microsoft} are mixed into 3D datasets for training. 

\begin{table*}[t]
\centering
\renewcommand\tabcolsep{3pt}
\caption{Comparison with state-of-the-art approaches on Human3.6M dataset. 
}
\vspace{-0.3cm}
\resizebox{.99\textwidth}{!}{
\begin{tabular}{lcccccccccccccccc}
\toprule
MPJPE(mm) & Direct. & Discuss & Eating & Greet & Phone & Photo & Pose & Purch. & Sitting & SitingD & Smoke & Wait & WalkD & Walk & WalkT & Avg\\
\midrule

Jahangiri~\etal~\cite{jahangiri2017generating} & 74.4 & 66.7 & 67.9 & 75.2 & 77.3 & 70.6 & 64.5 & 95.6 & 127.3 & 79.6 & 79.1 & 73.4 & 67.4 & 71.8 & 72.8 & 77.6\\
Mehta~\etal~\cite{mehta2017monocular} & 57.5 & 68.6 & 59.6 & 67.3 & 78.1 & 56.9 & 69.1 & 98.0 & 117.5 & 69.5 & 82.4 & 68.0 & 55.3 & 76.5 & 61.4 & 72.9\\
Martinez~\etal~\cite{martinez2017simple}  & 51.8 & 56.2 & 58.1 & 59.0 & 69.5 & 55.2 & 58.1 & 74.0 & 94.6 & 62.3 & 78.4 & 59.1 & 49.5 & 65.1 & 52.4 & 62.9\\
Sun~\etal~\cite{sun2017compositional}  & 52.8 & 54.8 & 54.2 & 54.3 & 61.8 & 53.1 & 53.6 & 71.7 & 86.7 & 61.5 & 67.2 & 53.4 & 47.1 & 61.6 & 63.4 & 59.1\\
Pavlakos~\etal~\cite{pavlakos2018ordinal}  & 48.5 & 54.4 & 54.4 & 52.0 & 59.4 & 65.3 & 49.9 & 52.9 & 65.8 & 71.1 & 56.6 & 52.9 & 60.9 & 44.7 & 47.8 & 56.2\\
Sun~\etal~\cite{sun2018integral}  & 47.5 & 47.7 & 49.5 & 50.2 & 51.4 & 43.8 & 46.4 & 58.9 & 65.7 & \textbf{49.4} & 55.8 & 47.8 & 38.9 & 49.0 & 43.8 & 49.6\\
Moon~\etal~\cite{moon2019camera}  & 50.5 & 55.7 & 50.1 & 51.7 & 53.9 & 46.8 & 50.0 & 61.9 & 68.0 & 52.5 & 55.9 & 49.9 & 41.8 & 56.1 & 46.9 & 53.3\\
Cai~\etal~\cite{cai2019exploiting} & 46.5 & 48.8 & 47.6 & 50.9 & 52.9 & 61.3 & 48.3 & 45.8 & 59.2 & 64.4 & 51.2 & 48.4 & 53.5 & 39.2 & 41.2 & 50.6\\
Zeng~\etal~\cite{zeng2020srnet} & - & - & - & - & - & - & - & - & - & - & - & - & - & - & - & 49.9\\
Wehrbein~\etal~\cite{wehrbein2021probabilistic} &38.5 & 42.6 & 39.9 & 41.7 & 46.5 & 51.6 & 39.9 & 40.8 & 49.5 & 56.8 & 45.3 & 46.4 & 46.8 & 37.8 & 40.4 & 44.3\\
Ma~\etal~\cite{ma2021context} & \textbf{36.3} & 42.8 & 39.5 & \textbf{40.0} & \textbf{43.9} & 48.8 & 36.7 & \textbf{44.0} & 51.0 & 63.1 & 44.3 & 40.6 & 44.4 & 34.9 & \textbf{36.7} & 43.4\\
\midrule
\textbf{Ours(GR-M3D)} & 37.1 & \textbf{40.4} &  \textbf{39.3} & 41.2 & \textbf{43.1} & \textbf{43.2} & \textbf{31.8} & 44.7 & \textbf{47.2} & 59.9 & \textbf{41.1} & \textbf{37.2} & \textbf{42.1} & \textbf{33.7} & 37.6 & \textbf{41.3} \\
\midrule
\midrule
PA MPJPE(mm) & Direct. & Discuss & Eating & Greet & Phone & Photo & Pose & Purch. & Sitting & SitingD & Smoke & Wait & WalkD & Walk & WalkT & Avg\\
\midrule
Martinez~\etal~\cite{martinez2017simple} & 39.5 & 43.2 & 46.4 & 47.0 & 51.0 & 41.4 & 40.6 & 56.5 & 69.4 & 49.2 & 56.0 & 45.0 & 38.0 & 49.5 & 43.1 & 47.7\\
Fang~\etal~\cite{fang2018learning} & 38.2 & 41.7 & 43.7 & 44.9 & 48.5 & 40.2 & 38.2 & 54.5 & 64.4 & 47.2 & 55.3 & 44.3 & 36.7 & 47.3 & 41.7 & 45.7\\
Sun ~\etal~\cite{sun2018integral} & 36.9 & 36.2 & 40.6 & 40.4 & 41.9 & 34.9 & 35.7 & 50.1 & 59.4 & 40.4 & 44.9 & 39.0 & 30.8 & 39.8 & 36.7 & 40.6\\
Cai~\etal~\cite{cai2019exploiting} & 36.8 & 38.7 & 38.2 & 41.7 & 40.7 & 46.8 & 37.9 & 35.6 & 47.6 & 51.7 & 41.3 & 36.8 & 42.7 & 31.0 & 34.7 & 40.2\\
Ma~\etal~\cite{ma2021context} & 30.5 & 34.9 & 32.0 & 32.2 & 35.0 & 37.8 & 28.6 & 32.6 & 40.8 & 52.0 & 35.0 & 31.9 & 35.6 & 26.6 & 28.5 & 34.6\\
Moon~\etal~\cite{moon2019camera} & 31.0 & 30.6 & 39.9 & 35.5 & 34.8 & \textbf{30.2} & 32.1 & 35.0 & 43.8 & \textbf{35.7} & 37.6 & 30.1 & 24.6 & 35.7 & 29.3 & 34.0\\
Wehrbein~\etal~\cite{wehrbein2021probabilistic} &27.9 & 31.4 & 29.7 & 30.2 & 34.9 & 37.1 & 27.3 & 28.2 & 39.0 & 46.1 & 34.2 & 32.3 & 33.6 & 26.1 & 27.5 & 32.4\\
\midrule
\textbf{Ours(GR-M3D)} & \textbf{24.4} & \textbf{26.3} & \textbf{25.4} & \textbf{26.5} & \textbf{30.6} & 31.4 & \textbf{24.3} & \textbf{29.7} &\textbf{ 30.2} & 36.4 & \textbf{27.5} & \textbf{23.4} & \textbf{22.9} &\textbf{ 24.8} & \textbf{25.2} & \textbf{27.3} \\
\bottomrule
\end{tabular}}

\label{table_sota_hm36_1}
\end{table*}



\begin{table}[t]
\centering
\renewcommand\tabcolsep{2pt}

\caption{Comparison on MuPoTS-3D, a 3D multi-person pose estimation dataset. ``-" shows that the results are not available. GR-M3D outperforms the SOTA methods. Bigger is better.}
\vspace{-0.3cm}
\resizebox{.99\columnwidth}{!}{
\begin{tabular}{l|c|ccc}
\toprule
 Methods  & Category & $PCK_{rel}$ & $PCK_{abs}$ & $AUC_{rel}$\\

\midrule
Lcr-net~\cite{rogez2019lcr} & \multirow{5}{*}{Top-down} & 62.4 & - & - \\
Lcr-net++~\cite{rogez2019lcr++} &  & 74.0 & - & - \\
HG-RCNN~\cite{dabral2019multi} &   & 74.2 & - & - \\
HMOR~\cite{wang2020hmor} &  & 82.0 & - & - \\
PoseNet~\cite{moon2019camera} &  & \textbf{82.5} & \textbf{31.8} & \textbf{40.9} \\
\midrule
ORPM~\cite{mehta2018single} & \multirow{3}{*}{Bottom-up} & 69.8 & - & -  \\
Xnect~\cite{mehta2019xnect} &  & 75.8 & - & - \\
SMAP~\cite{zhen2020smap} &  & \textbf{80.5} & \textbf{38.7} & \textbf{42.7} \\
\midrule
\textbf{Ours(GR-M3D)} & Bottom-up & \textbf{84.6}$\uparrow$2.5\% & \textbf{41.2}$\uparrow$6.5\% & \textbf{44.1}$\uparrow$3.3\% \\
\bottomrule
\end{tabular}
}

\label{table_multi_person}
\end{table}

\begin{table}[]
\centering
\renewcommand\tabcolsep{5pt}

\caption{Comparison with state-of-the-art methods on CMU Panoptic dataset, 3D multi-person pose estimation dataset. $*$ denotes refining results by an extra network. Lower is better.}
\label{table_cmu}
\vspace{-0.3cm}

\resizebox{.99\columnwidth}{!}{
\begin{tabular}{l|cccc|c}
\toprule
 MPJPE(mm) & Hagg. & Mafia & Ultim. & Pizza & Avg. \\
\midrule
 DMHS~\cite{popa2017deep} & 217.9 & 187.3 & 193.6 & 221.3 & 203.4\\
 SemanticFB~\cite{zanfir2018monocular} & 140.0 & 165.9 & 150.7 & 156.0 & 153.4\\
 PoseNet~\cite{moon2019camera} & 89.6 & 91.3 & 79.6 & 90.1 & 87.6\\
 MubyNet~\cite{zanfir2018deep} & 72.4 & 78.8 & 66.8 & 94.3 & 78.1\\
 SMAP~\cite{zhen2020smap} & 71.8 & 72.5 & 65.9 & 82.1 & 73.1\\
 LoCO~\cite{fabbri2020compressed} & 45.0 & 95.0 & 58.0 & 79.0 & 69.0\\
 SMAP~\cite{zhen2020smap}$^*$ & 63.1 & 60.3 & 56.6 & 67.1 & 61.8\\
 \midrule
\textbf{Ours(GR-M3D)} & \textbf{57.1} & \textbf{58.3} & \textbf{53.4} & \textbf{62.7} & \textbf{57.9$\downarrow 6.3\%$} \\
\bottomrule
\end{tabular}
}
\end{table}

\subsection{Ablation study}
Our baseline is the recently-developed offset-based method. They learn keypoint 3D offset and then decode human pose from single root keypoint by the corresponding star or tree pose graphs~\cite{nie2019single,zhou2019objects}.

\textbf{Effectiveness of SDAR and DGR}
For the ablation study, we use ResNet-34 as the backbone. The ablation study of SDAR and DGR is shown in table \ref{table_ablation_modules}. 
We can observe that decoding the 3D pose with the tree graph is better than the star graph, which gains a relative improvement of 1.7\%.
Our DGR achieves 77.9 3DPCK, which gains a relative improvement of 3.3\%, compared with the star graph decoding method. We further use the scale and depth aware refinement module to improve the learned keypoints heatmaps, scale maps, depth maps, and 3D offset maps. SDAR achieves 78.6 3DPCK based on our DGR decoding approach. The whole gain of GR-M3D with SDAR and DGR is 3.2 3DPCK, a relative improvement of 4.2\%, compared with the star structure decoding approach~\cite{nie2019single}.

We also make a comparison of star, tree, and DGR decoding in Figure \ref{fig:case_ana}. As shown in Figure \ref{fig:case_ana}, the decoding methods based on star and tree structure fail in occlusion cases. Our DGR performs well on these challenging cases since the multi-root-based dynamic graph reasoning mechanism.

\begin{figure*}[t]
\centering
\includegraphics[width=0.95\textwidth]{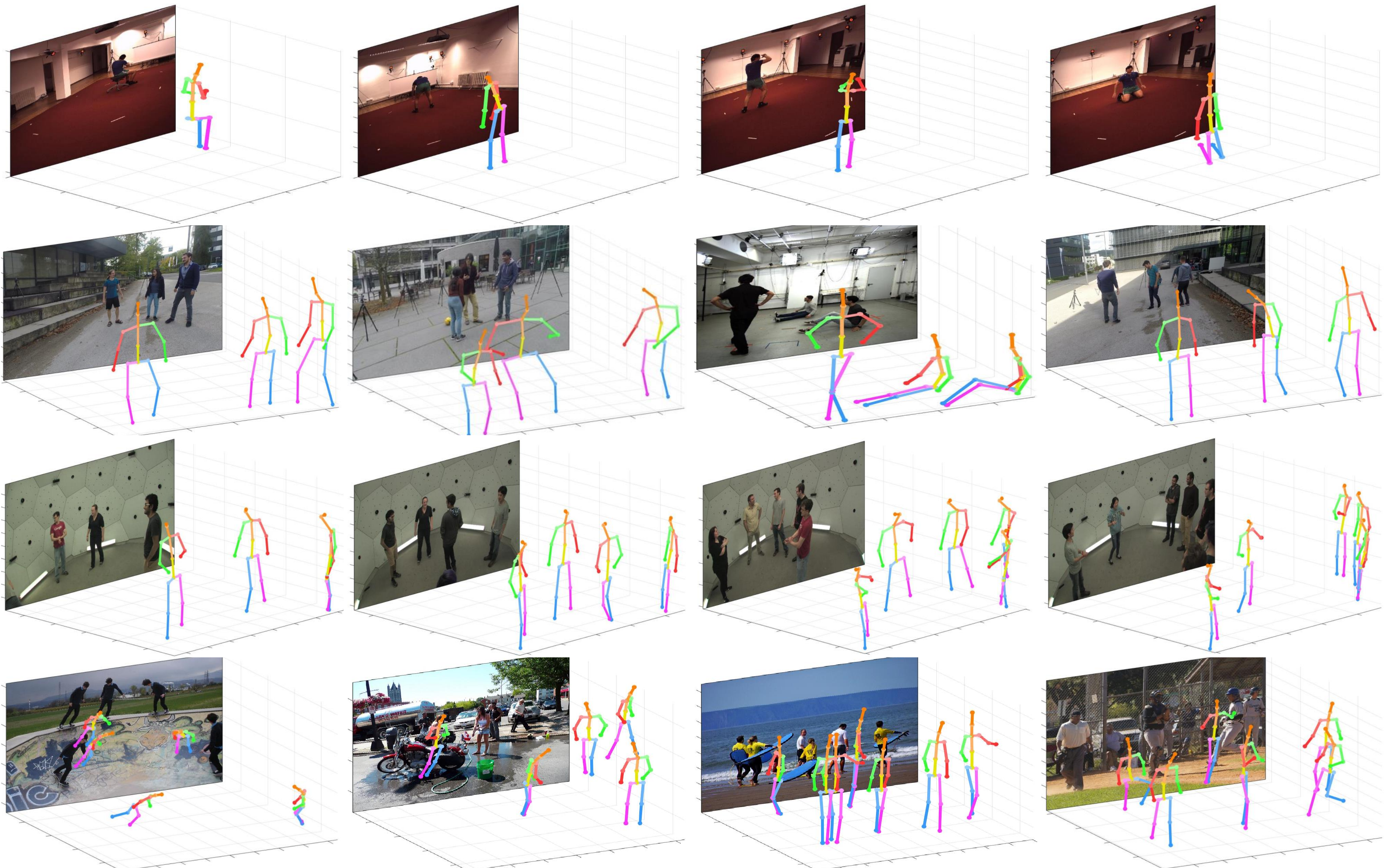} 
\vspace{-0.2cm}
\caption{Visualization of the predicted 3D poses by GR-M3D. 
}
\label{case_vis}
\end{figure*}

\textbf{Different backbones}
To evaluate the effectiveness of SDAR and DGR on different backbones, 
we use several widely used networks ( Hourglass~\cite{newell2016stacked} , ResNet~\cite{he2016deep}, HRNet-32~\cite{sun2019deep}) as backbone. 
As shown in table \ref{table_backbones}, 
our GR-M3D outperforms baseline over a relative of 3\% in different backbones.
GR-M3D achieves 84.6 3DPCK on the MuPoTS-3D dataset, which is based on the Hourglass backbone.
GR-M3D is lightweight and fast. 
GR-M3D only increases 2MB parameters, compared with baseline methods. GR-M3D achieves real-time based on ResNet-34 and ResNet-50. 
Even with the heavy backbones (ResNet-101, HRNet-32), GR-M3D achieves 15-23 fps, which indicates that GR-M3D has broad application potential.

\textbf{Evaluation in crowded scenes}
To evaluate the ability of GR-M3D to handle the occlusion cases, 
which are often seen in the crowded scenes, 
we evaluate GR-M3D on MuPoTS-3D with different \textit{Crowd Index}. 
The bigger value of \textit{ Index} means more crowded cases in the image. 
That represents the pose variances and occlusions are more serious. 
As shown in table \ref{table_crowd_index}, with the increase of \textit{Crowd Index}, the 3DPCK of baseline decreases. 
At the same time, our GR-M3D outperforms baseline in all \textit{Crowd Index}. 
Even in the most challenging cases, with a crowd index of over 0.7, GR-M3D outperforms baseline with a relative improvement of 5.9\%. 
These results show that our GR-M3D can handle the problems of pose variances and heavy occlusions.

\subsection{Comparison with state-of-the-art methods}

Following the settings of SOTA methods, we report the results of GR-M3D on the Human3.6M dataset. As shown in table \ref{table_sota_hm36_1}, GR-M3D outperforms SOTA methods and achieves gains of 5\% and 16\% in MPJPE and PA MPJPE based on Hourglass backbone, respectively. 

For multi-person cases, the commonly used dataset is the MuCo-3DHP and MuPoTS-3D dataset. We train GR-M3D on the MuCo-3DHP and test on the MuPoTS-3D. The results are shown in table \ref{table_multi_person}. Based on Hourglass backbone, GR-M3D achieves the state-of-the-art results and obtains relative improvements of 2.5\%, 6.5\%, and 3.3\% in $PCK_{rel}$, $PCK_{abs}$, and $AUC_{rel}$, respectively. 

CMU Panoptic dataset is another widely-used benchmark for multi-person 3D pose estimation. Following \cite{zanfir2018deep,zhen2020smap}, we take experiments on this dataset and show results in table \ref{table_cmu}. Compared with state-of-the-art methods, GR-M3D obtains a relative improvement of 6.3\% in MPJPE based on the Hourglass backbone.

 

\subsection{Generalization in the wild}
The images in MuPoTS-3D are collected in the wild scenes. Table \ref{table_multi_person} has shown that GR-M3D outperforms the state-of-the-art methods, which demonstrate the generalization ability of GR-M3D in the wild scenes. 
We also conduct experiments on COCO~\cite{lin2014microsoft}, a larger scale 2D pose estimation dataset. All of the images are collected in challenging, uncontrolled conditions. We directly predict the 3D pose on the COCO images since no 3D pose annotations. The model is based on Hourglass and trained on  MuCo-3DHP. 
The results are shown in Figure \ref{case_vis}. 
The visualization results in Figure \ref{case_vis} are from COCO, Human3.6M, MuPoTS-3D, and CMU Panoptic datasets, respectively. 
GR-M3D performs well on these challenging cases, even on the images collected in the outdoor scenes in the COCO dataset, which shows the generalization ability of GR-M3D.

\section{Conclusion}
We propose a novel bottom-up approach (GR-M3D) for 3D multi-person pose estimation, which mitigates occlusions and depth ambiguity by capturing more global context information. We firstly design a scale and depth refinement (SDAR) module to enhance the learned feature maps, to further generate better root keypoints and build robust message propagation paths. DGR reasons the dynamic decoding graphs from the predicted message propagation paths to decoding 3D poses. GR-M3D outperforms previous works and achieves state-of-the-art results on three widely-used benchmarks.

\section{Acknowledgement}
This work was supported by the Scientific and Technological Innovation of Shunde Graduate School of University of Science and Technology Beijing (No. BK20AE004 and No.BK19CE017).

\clearpage
\bibliographystyle{ACM-Reference-Format}
\bibliography{sample-base}


\begin{thebibliography}{55}


\ifx \showCODEN    \undefined \def \showCODEN     #1{\unskip}     \fi
\ifx \showDOI      \undefined \def \showDOI       #1{#1}\fi
\ifx \showISBNx    \undefined \def \showISBNx     #1{\unskip}     \fi
\ifx \showISBNxiii \undefined \def \showISBNxiii  #1{\unskip}     \fi
\ifx \showISSN     \undefined \def \showISSN      #1{\unskip}     \fi
\ifx \showLCCN     \undefined \def \showLCCN      #1{\unskip}     \fi
\ifx \shownote     \undefined \def \shownote      #1{#1}          \fi
\ifx \showarticletitle \undefined \def \showarticletitle #1{#1}   \fi
\ifx \showURL      \undefined \def \showURL       {\relax}        \fi
\providecommand\bibfield[2]{#2}
\providecommand\bibinfo[2]{#2}
\providecommand\natexlab[1]{#1}
\providecommand\showeprint[2][]{arXiv:#2}

\bibitem[\protect\citeauthoryear{Andriluka, Pishchulin, Gehler, and
  Schiele}{Andriluka et~al\mbox{.}}{2014}]%
        {andriluka20142d}
\bibfield{author}{\bibinfo{person}{Mykhaylo Andriluka}, \bibinfo{person}{Leonid
  Pishchulin}, \bibinfo{person}{Peter Gehler}, {and} \bibinfo{person}{Bernt
  Schiele}.} \bibinfo{year}{2014}\natexlab{}.
\newblock \showarticletitle{2d human pose estimation: New benchmark and state
  of the art analysis}. In \bibinfo{booktitle}{\emph{CVPR}}.
  \bibinfo{pages}{3686--3693}.
\newblock


\bibitem[\protect\citeauthoryear{Cai, Ge, Liu, Cai, Cham, Yuan, and
  Thalmann}{Cai et~al\mbox{.}}{2019}]%
        {cai2019exploiting}
\bibfield{author}{\bibinfo{person}{Yujun Cai}, \bibinfo{person}{Liuhao Ge},
  \bibinfo{person}{Jun Liu}, \bibinfo{person}{Jianfei Cai},
  \bibinfo{person}{Tat-Jen Cham}, \bibinfo{person}{Junsong Yuan}, {and}
  \bibinfo{person}{Nadia~Magnenat Thalmann}.} \bibinfo{year}{2019}\natexlab{}.
\newblock \showarticletitle{Exploiting spatial-temporal relationships for 3d
  pose estimation via graph convolutional networks}. In
  \bibinfo{booktitle}{\emph{ICCV}}. \bibinfo{pages}{2272--2281}.
\newblock


\bibitem[\protect\citeauthoryear{Cao, Simon, Wei, and Sheikh}{Cao
  et~al\mbox{.}}{2017}]%
        {cao2017realtime}
\bibfield{author}{\bibinfo{person}{Zhe Cao}, \bibinfo{person}{Tomas Simon},
  \bibinfo{person}{Shih-En Wei}, {and} \bibinfo{person}{Yaser Sheikh}.}
  \bibinfo{year}{2017}\natexlab{}.
\newblock \showarticletitle{Realtime multi-person 2d pose estimation using part
  affinity fields}. In \bibinfo{booktitle}{\emph{CVPR}}.
  \bibinfo{pages}{7291--7299}.
\newblock


\bibitem[\protect\citeauthoryear{Chen, Wang, Peng, Zhang, Yu, and Sun}{Chen
  et~al\mbox{.}}{2018}]%
        {chen2018cascaded}
\bibfield{author}{\bibinfo{person}{Yilun Chen}, \bibinfo{person}{Zhicheng
  Wang}, \bibinfo{person}{Yuxiang Peng}, \bibinfo{person}{Zhiqiang Zhang},
  \bibinfo{person}{Gang Yu}, {and} \bibinfo{person}{Jian Sun}.}
  \bibinfo{year}{2018}\natexlab{}.
\newblock \showarticletitle{Cascaded pyramid network for multi-person pose
  estimation}. In \bibinfo{booktitle}{\emph{CVPR}}.
  \bibinfo{pages}{7103--7112}.
\newblock


\bibitem[\protect\citeauthoryear{Cheng, Wang, Yang, and Tan}{Cheng
  et~al\mbox{.}}{2021}]%
        {cheng2021monocular}
\bibfield{author}{\bibinfo{person}{Yu Cheng}, \bibinfo{person}{Bo Wang},
  \bibinfo{person}{Bo Yang}, {and} \bibinfo{person}{Robby~T Tan}.}
  \bibinfo{year}{2021}\natexlab{}.
\newblock \showarticletitle{Monocular 3D Multi-Person Pose Estimation by
  Integrating Top-Down and Bottom-Up Networks}. In
  \bibinfo{booktitle}{\emph{CVPR}}. \bibinfo{pages}{7649--7659}.
\newblock


\bibitem[\protect\citeauthoryear{Ci, Wang, Ma, and Wang}{Ci
  et~al\mbox{.}}{2019}]%
        {ci2019optimizing}
\bibfield{author}{\bibinfo{person}{Hai Ci}, \bibinfo{person}{Chunyu Wang},
  \bibinfo{person}{Xiaoxuan Ma}, {and} \bibinfo{person}{Yizhou Wang}.}
  \bibinfo{year}{2019}\natexlab{}.
\newblock \showarticletitle{Optimizing network structure for 3d human pose
  estimation}. In \bibinfo{booktitle}{\emph{ICCV}}.
  \bibinfo{pages}{2262--2271}.
\newblock


\bibitem[\protect\citeauthoryear{Dabral, Gundavarapu, Mitra, Sharma,
  Ramakrishnan, and Jain}{Dabral et~al\mbox{.}}{2019}]%
        {dabral2019multi}
\bibfield{author}{\bibinfo{person}{Rishabh Dabral}, \bibinfo{person}{Nitesh~B
  Gundavarapu}, \bibinfo{person}{Rahul Mitra}, \bibinfo{person}{Abhishek
  Sharma}, \bibinfo{person}{Ganesh Ramakrishnan}, {and} \bibinfo{person}{Arjun
  Jain}.} \bibinfo{year}{2019}\natexlab{}.
\newblock \showarticletitle{Multi-person 3d human pose estimation from
  monocular images}. In \bibinfo{booktitle}{\emph{3DV}}. IEEE,
  \bibinfo{pages}{405--414}.
\newblock


\bibitem[\protect\citeauthoryear{Eigen, Puhrsch, and Fergus}{Eigen
  et~al\mbox{.}}{2014}]%
        {NIPS2014_7bccfde7}
\bibfield{author}{\bibinfo{person}{David Eigen}, \bibinfo{person}{Christian
  Puhrsch}, {and} \bibinfo{person}{Rob Fergus}.}
  \bibinfo{year}{2014}\natexlab{}.
\newblock \showarticletitle{Depth Map Prediction from a Single Image using a
  Multi-Scale Deep Network}. In \bibinfo{booktitle}{\emph{NeurIPS}},
  \bibfield{editor}{\bibinfo{person}{Z.~Ghahramani},
  \bibinfo{person}{M.~Welling}, \bibinfo{person}{C.~Cortes},
  \bibinfo{person}{N.~Lawrence}, {and} \bibinfo{person}{K.~Q. Weinberger}}
  (Eds.), Vol.~\bibinfo{volume}{27}. \bibinfo{publisher}{Curran Associates,
  Inc.}
\newblock
\urldef\tempurl%
\url{https://proceedings.neurips.cc/paper/2014/file/7bccfde7714a1ebadf06c5f4cea752c1-Paper.pdf}
\showURL{%
\tempurl}


\bibitem[\protect\citeauthoryear{Fabbri, Lanzi, Calderara, Alletto, and
  Cucchiara}{Fabbri et~al\mbox{.}}{2020}]%
        {fabbri2020compressed}
\bibfield{author}{\bibinfo{person}{Matteo Fabbri}, \bibinfo{person}{Fabio
  Lanzi}, \bibinfo{person}{Simone Calderara}, \bibinfo{person}{Stefano
  Alletto}, {and} \bibinfo{person}{Rita Cucchiara}.}
  \bibinfo{year}{2020}\natexlab{}.
\newblock \showarticletitle{Compressed volumetric heatmaps for multi-person 3d
  pose estimation}. In \bibinfo{booktitle}{\emph{CVPR}}.
  \bibinfo{pages}{7204--7213}.
\newblock


\bibitem[\protect\citeauthoryear{Fang, Xu, Wang, Liu, and Zhu}{Fang
  et~al\mbox{.}}{2018}]%
        {fang2018learning}
\bibfield{author}{\bibinfo{person}{Hao-Shu Fang}, \bibinfo{person}{Yuanlu Xu},
  \bibinfo{person}{Wenguan Wang}, \bibinfo{person}{Xiaobai Liu}, {and}
  \bibinfo{person}{Song-Chun Zhu}.} \bibinfo{year}{2018}\natexlab{}.
\newblock \showarticletitle{Learning pose grammar to encode human body
  configuration for 3d pose estimation}. In \bibinfo{booktitle}{\emph{AAAI}},
  Vol.~\bibinfo{volume}{32}.
\newblock


\bibitem[\protect\citeauthoryear{Gower}{Gower}{1975}]%
        {gower1975generalized}
\bibfield{author}{\bibinfo{person}{John~C Gower}.}
  \bibinfo{year}{1975}\natexlab{}.
\newblock \showarticletitle{Generalized procrustes analysis}.
\newblock \bibinfo{journal}{\emph{Psychometrika}} \bibinfo{volume}{40},
  \bibinfo{number}{1} (\bibinfo{year}{1975}), \bibinfo{pages}{33--51}.
\newblock


\bibitem[\protect\citeauthoryear{He, Zhang, Ren, and Sun}{He
  et~al\mbox{.}}{2016}]%
        {he2016deep}
\bibfield{author}{\bibinfo{person}{Kaiming He}, \bibinfo{person}{Xiangyu
  Zhang}, \bibinfo{person}{Shaoqing Ren}, {and} \bibinfo{person}{Jian Sun}.}
  \bibinfo{year}{2016}\natexlab{}.
\newblock \showarticletitle{Deep residual learning for image recognition}. In
  \bibinfo{booktitle}{\emph{CVPR}}. \bibinfo{pages}{770--778}.
\newblock


\bibitem[\protect\citeauthoryear{Ionescu, Papava, Olaru, and
  Sminchisescu}{Ionescu et~al\mbox{.}}{2013}]%
        {ionescu2013human3}
\bibfield{author}{\bibinfo{person}{Catalin Ionescu}, \bibinfo{person}{Dragos
  Papava}, \bibinfo{person}{Vlad Olaru}, {and} \bibinfo{person}{Cristian
  Sminchisescu}.} \bibinfo{year}{2013}\natexlab{}.
\newblock \showarticletitle{Human3.6m: Large scale datasets and predictive
  methods for 3d human sensing in natural environments}.
\newblock \bibinfo{journal}{\emph{TPAMI}} \bibinfo{volume}{36},
  \bibinfo{number}{7} (\bibinfo{year}{2013}), \bibinfo{pages}{1325--1339}.
\newblock


\bibitem[\protect\citeauthoryear{Jahangiri and Yuille}{Jahangiri and
  Yuille}{2017}]%
        {jahangiri2017generating}
\bibfield{author}{\bibinfo{person}{Ehsan Jahangiri} {and}
  \bibinfo{person}{Alan~L Yuille}.} \bibinfo{year}{2017}\natexlab{}.
\newblock \showarticletitle{Generating multiple diverse hypotheses for human 3d
  pose consistent with 2d joint detections}. In
  \bibinfo{booktitle}{\emph{CVPRW}}. \bibinfo{pages}{805--814}.
\newblock


\bibitem[\protect\citeauthoryear{Joo, Simon, Li, Liu, Tan, Gui, Banerjee,
  Godisart, Nabbe, Matthews, et~al\mbox{.}}{Joo et~al\mbox{.}}{2017}]%
        {joo2017panoptic}
\bibfield{author}{\bibinfo{person}{Hanbyul Joo}, \bibinfo{person}{Tomas Simon},
  \bibinfo{person}{Xulong Li}, \bibinfo{person}{Hao Liu}, \bibinfo{person}{Lei
  Tan}, \bibinfo{person}{Lin Gui}, \bibinfo{person}{Sean Banerjee},
  \bibinfo{person}{Timothy Godisart}, \bibinfo{person}{Bart Nabbe},
  \bibinfo{person}{Iain Matthews}, {et~al\mbox{.}}}
  \bibinfo{year}{2017}\natexlab{}.
\newblock \showarticletitle{Panoptic studio: A massively multiview system for
  social interaction capture}.
\newblock \bibinfo{journal}{\emph{TPAMI}} \bibinfo{volume}{41},
  \bibinfo{number}{1} (\bibinfo{year}{2017}), \bibinfo{pages}{190--204}.
\newblock


\bibitem[\protect\citeauthoryear{Kanazawa, Black, Jacobs, and Malik}{Kanazawa
  et~al\mbox{.}}{2018}]%
        {kanazawa2018end}
\bibfield{author}{\bibinfo{person}{Angjoo Kanazawa}, \bibinfo{person}{Michael~J
  Black}, \bibinfo{person}{David~W Jacobs}, {and} \bibinfo{person}{Jitendra
  Malik}.} \bibinfo{year}{2018}\natexlab{}.
\newblock \showarticletitle{End-to-end recovery of human shape and pose}. In
  \bibinfo{booktitle}{\emph{CVPR}}. \bibinfo{pages}{7122--7131}.
\newblock


\bibitem[\protect\citeauthoryear{Kocabas, Karagoz, and Akbas}{Kocabas
  et~al\mbox{.}}{2018}]%
        {kocabas2018multiposenet}
\bibfield{author}{\bibinfo{person}{Muhammed Kocabas}, \bibinfo{person}{Salih
  Karagoz}, {and} \bibinfo{person}{Emre Akbas}.}
  \bibinfo{year}{2018}\natexlab{}.
\newblock \showarticletitle{Multiposenet: Fast multi-person pose estimation
  using pose residual network}. In \bibinfo{booktitle}{\emph{ECCV}}.
  \bibinfo{pages}{417--433}.
\newblock


\bibitem[\protect\citeauthoryear{Kreiss, Bertoni, and Alahi}{Kreiss
  et~al\mbox{.}}{2019}]%
        {kreiss2019pifpaf}
\bibfield{author}{\bibinfo{person}{Sven Kreiss}, \bibinfo{person}{Lorenzo
  Bertoni}, {and} \bibinfo{person}{Alexandre Alahi}.}
  \bibinfo{year}{2019}\natexlab{}.
\newblock \showarticletitle{Pifpaf: Composite fields for human pose
  estimation}. In \bibinfo{booktitle}{\emph{CVPR}}.
  \bibinfo{pages}{11977--11986}.
\newblock


\bibitem[\protect\citeauthoryear{Li, Wang, Zhu, Mao, Fang, and Lu}{Li
  et~al\mbox{.}}{2019}]%
        {li2019crowdpose}
\bibfield{author}{\bibinfo{person}{Jiefeng Li}, \bibinfo{person}{Can Wang},
  \bibinfo{person}{Hao Zhu}, \bibinfo{person}{Yihuan Mao},
  \bibinfo{person}{Hao-Shu Fang}, {and} \bibinfo{person}{Cewu Lu}.}
  \bibinfo{year}{2019}\natexlab{}.
\newblock \showarticletitle{Crowdpose: Efficient crowded scenes pose estimation
  and a new benchmark}. In \bibinfo{booktitle}{\emph{CVPR}}.
  \bibinfo{pages}{10863--10872}.
\newblock


\bibitem[\protect\citeauthoryear{Li, Liu, Lu, Wang, Liu, Li, and Lu}{Li
  et~al\mbox{.}}{2020}]%
        {li2020detailed}
\bibfield{author}{\bibinfo{person}{Yong-Lu Li}, \bibinfo{person}{Xinpeng Liu},
  \bibinfo{person}{Han Lu}, \bibinfo{person}{Shiyi Wang},
  \bibinfo{person}{Junqi Liu}, \bibinfo{person}{Jiefeng Li}, {and}
  \bibinfo{person}{Cewu Lu}.} \bibinfo{year}{2020}\natexlab{}.
\newblock \showarticletitle{Detailed 2d-3d joint representation for
  human-object interaction}. In \bibinfo{booktitle}{\emph{CVPR}}.
  \bibinfo{pages}{10166--10175}.
\newblock


\bibitem[\protect\citeauthoryear{Lin, Maire, Belongie, Hays, Perona, Ramanan,
  Doll{\'a}r, and Zitnick}{Lin et~al\mbox{.}}{2014}]%
        {lin2014microsoft}
\bibfield{author}{\bibinfo{person}{Tsung-Yi Lin}, \bibinfo{person}{Michael
  Maire}, \bibinfo{person}{Serge Belongie}, \bibinfo{person}{James Hays},
  \bibinfo{person}{Pietro Perona}, \bibinfo{person}{Deva Ramanan},
  \bibinfo{person}{Piotr Doll{\'a}r}, {and} \bibinfo{person}{C~Lawrence
  Zitnick}.} \bibinfo{year}{2014}\natexlab{}.
\newblock \showarticletitle{Microsoft coco: Common objects in context}. In
  \bibinfo{booktitle}{\emph{ECCV}}. Springer, \bibinfo{pages}{740--755}.
\newblock


\bibitem[\protect\citeauthoryear{Liu, Zhang, Chen, Wang, and Ouyang}{Liu
  et~al\mbox{.}}{2020}]%
        {liu2020disentangling}
\bibfield{author}{\bibinfo{person}{Ziyu Liu}, \bibinfo{person}{Hongwen Zhang},
  \bibinfo{person}{Zhenghao Chen}, \bibinfo{person}{Zhiyong Wang}, {and}
  \bibinfo{person}{Wanli Ouyang}.} \bibinfo{year}{2020}\natexlab{}.
\newblock \showarticletitle{Disentangling and unifying graph convolutions for
  skeleton-based action recognition}. In \bibinfo{booktitle}{\emph{CVPR}}.
  \bibinfo{pages}{143--152}.
\newblock


\bibitem[\protect\citeauthoryear{Ma, Su, Wang, Ci, and Wang}{Ma
  et~al\mbox{.}}{2021}]%
        {ma2021context}
\bibfield{author}{\bibinfo{person}{Xiaoxuan Ma}, \bibinfo{person}{Jiajun Su},
  \bibinfo{person}{Chunyu Wang}, \bibinfo{person}{Hai Ci}, {and}
  \bibinfo{person}{Yizhou Wang}.} \bibinfo{year}{2021}\natexlab{}.
\newblock \showarticletitle{Context Modeling in 3D Human Pose Estimation: A
  Unified Perspective}. In \bibinfo{booktitle}{\emph{CVPR}}.
  \bibinfo{pages}{6238--6247}.
\newblock


\bibitem[\protect\citeauthoryear{Martinez, Hossain, Romero, and
  Little}{Martinez et~al\mbox{.}}{2017}]%
        {martinez2017simple}
\bibfield{author}{\bibinfo{person}{Julieta Martinez}, \bibinfo{person}{Rayat
  Hossain}, \bibinfo{person}{Javier Romero}, {and} \bibinfo{person}{James~J
  Little}.} \bibinfo{year}{2017}\natexlab{}.
\newblock \showarticletitle{A simple yet effective baseline for 3d human pose
  estimation}. In \bibinfo{booktitle}{\emph{ICCV}}.
  \bibinfo{pages}{2640--2649}.
\newblock


\bibitem[\protect\citeauthoryear{Mehta, Rhodin, Casas, Fua, Sotnychenko, Xu,
  and Theobalt}{Mehta et~al\mbox{.}}{2017}]%
        {mehta2017monocular}
\bibfield{author}{\bibinfo{person}{Dushyant Mehta}, \bibinfo{person}{Helge
  Rhodin}, \bibinfo{person}{Dan Casas}, \bibinfo{person}{Pascal Fua},
  \bibinfo{person}{Oleksandr Sotnychenko}, \bibinfo{person}{Weipeng Xu}, {and}
  \bibinfo{person}{Christian Theobalt}.} \bibinfo{year}{2017}\natexlab{}.
\newblock \showarticletitle{Monocular 3d human pose estimation in the wild
  using improved cnn supervision}. In \bibinfo{booktitle}{\emph{3DV}}. IEEE,
  \bibinfo{pages}{506--516}.
\newblock


\bibitem[\protect\citeauthoryear{Mehta, Sotnychenko, Mueller, Xu, Elgharib,
  Fua, Seidel, Rhodin, Pons-Moll, and Theobalt}{Mehta et~al\mbox{.}}{2019}]%
        {mehta2019xnect}
\bibfield{author}{\bibinfo{person}{Dushyant Mehta}, \bibinfo{person}{Oleksandr
  Sotnychenko}, \bibinfo{person}{Franziska Mueller}, \bibinfo{person}{Weipeng
  Xu}, \bibinfo{person}{Mohamed Elgharib}, \bibinfo{person}{Pascal Fua},
  \bibinfo{person}{Hans-Peter Seidel}, \bibinfo{person}{Helge Rhodin},
  \bibinfo{person}{Gerard Pons-Moll}, {and} \bibinfo{person}{Christian
  Theobalt}.} \bibinfo{year}{2019}\natexlab{}.
\newblock \showarticletitle{Xnect: Real-time multi-person 3d human pose
  estimation with a single rgb camera}.
\newblock \bibinfo{journal}{\emph{arXiv preprint arXiv:1907.00837}}
  (\bibinfo{year}{2019}).
\newblock


\bibitem[\protect\citeauthoryear{Mehta, Sotnychenko, Mueller, Xu, Sridhar,
  Pons-Moll, and Theobalt}{Mehta et~al\mbox{.}}{2018}]%
        {mehta2018single}
\bibfield{author}{\bibinfo{person}{Dushyant Mehta}, \bibinfo{person}{Oleksandr
  Sotnychenko}, \bibinfo{person}{Franziska Mueller}, \bibinfo{person}{Weipeng
  Xu}, \bibinfo{person}{Srinath Sridhar}, \bibinfo{person}{Gerard Pons-Moll},
  {and} \bibinfo{person}{Christian Theobalt}.} \bibinfo{year}{2018}\natexlab{}.
\newblock \showarticletitle{Single-shot multi-person 3d pose estimation from
  monocular rgb}. In \bibinfo{booktitle}{\emph{3DV}}. IEEE,
  \bibinfo{pages}{120--130}.
\newblock


\bibitem[\protect\citeauthoryear{Moon, Chang, and Lee}{Moon
  et~al\mbox{.}}{2019}]%
        {moon2019camera}
\bibfield{author}{\bibinfo{person}{Gyeongsik Moon}, \bibinfo{person}{Ju~Yong
  Chang}, {and} \bibinfo{person}{Kyoung~Mu Lee}.}
  \bibinfo{year}{2019}\natexlab{}.
\newblock \showarticletitle{Camera distance-aware top-down approach for 3d
  multi-person pose estimation from a single rgb image}. In
  \bibinfo{booktitle}{\emph{ICCV}}. \bibinfo{pages}{10133--10142}.
\newblock


\bibitem[\protect\citeauthoryear{Moreno-Noguer}{Moreno-Noguer}{2017}]%
        {moreno20173d}
\bibfield{author}{\bibinfo{person}{Francesc Moreno-Noguer}.}
  \bibinfo{year}{2017}\natexlab{}.
\newblock \showarticletitle{3d human pose estimation from a single image via
  distance matrix regression}. In \bibinfo{booktitle}{\emph{CVPR}}.
  \bibinfo{pages}{2823--2832}.
\newblock


\bibitem[\protect\citeauthoryear{Newell, Yang, and Deng}{Newell
  et~al\mbox{.}}{2016}]%
        {newell2016stacked}
\bibfield{author}{\bibinfo{person}{Alejandro Newell}, \bibinfo{person}{Kaiyu
  Yang}, {and} \bibinfo{person}{Jia Deng}.} \bibinfo{year}{2016}\natexlab{}.
\newblock \showarticletitle{Stacked hourglass networks for human pose
  estimation}. In \bibinfo{booktitle}{\emph{ECCV}}. Springer,
  \bibinfo{pages}{483--499}.
\newblock


\bibitem[\protect\citeauthoryear{Nie, Feng, Zhang, and Yan}{Nie
  et~al\mbox{.}}{2019}]%
        {nie2019single}
\bibfield{author}{\bibinfo{person}{Xuecheng Nie}, \bibinfo{person}{Jiashi
  Feng}, \bibinfo{person}{Jianfeng Zhang}, {and} \bibinfo{person}{Shuicheng
  Yan}.} \bibinfo{year}{2019}\natexlab{}.
\newblock \showarticletitle{Single-stage multi-person pose machines}. In
  \bibinfo{booktitle}{\emph{ICCV}}. \bibinfo{pages}{6951--6960}.
\newblock


\bibitem[\protect\citeauthoryear{Pavlakos, Zhou, and Daniilidis}{Pavlakos
  et~al\mbox{.}}{2018}]%
        {pavlakos2018ordinal}
\bibfield{author}{\bibinfo{person}{Georgios Pavlakos}, \bibinfo{person}{Xiaowei
  Zhou}, {and} \bibinfo{person}{Kostas Daniilidis}.}
  \bibinfo{year}{2018}\natexlab{}.
\newblock \showarticletitle{Ordinal depth supervision for 3d human pose
  estimation}. In \bibinfo{booktitle}{\emph{CVPR}}.
  \bibinfo{pages}{7307--7316}.
\newblock


\bibitem[\protect\citeauthoryear{Pavlakos, Zhou, Derpanis, and
  Daniilidis}{Pavlakos et~al\mbox{.}}{2017}]%
        {pavlakos2017coarse}
\bibfield{author}{\bibinfo{person}{Georgios Pavlakos}, \bibinfo{person}{Xiaowei
  Zhou}, \bibinfo{person}{Konstantinos~G Derpanis}, {and}
  \bibinfo{person}{Kostas Daniilidis}.} \bibinfo{year}{2017}\natexlab{}.
\newblock \showarticletitle{Coarse-to-fine volumetric prediction for
  single-image 3D human pose}. In \bibinfo{booktitle}{\emph{CVPR}}.
  \bibinfo{pages}{7025--7034}.
\newblock


\bibitem[\protect\citeauthoryear{Popa, Zanfir, and Sminchisescu}{Popa
  et~al\mbox{.}}{2017}]%
        {popa2017deep}
\bibfield{author}{\bibinfo{person}{Alin-Ionut Popa}, \bibinfo{person}{Mihai
  Zanfir}, {and} \bibinfo{person}{Cristian Sminchisescu}.}
  \bibinfo{year}{2017}\natexlab{}.
\newblock \showarticletitle{Deep multitask architecture for integrated 2d and
  3d human sensing}. In \bibinfo{booktitle}{\emph{CVPR}}.
  \bibinfo{pages}{6289--6298}.
\newblock


\bibitem[\protect\citeauthoryear{Qiu, Qiu, Fu, and Fu}{Qiu
  et~al\mbox{.}}{2019}]%
        {qiu2019learning}
\bibfield{author}{\bibinfo{person}{Zhongwei Qiu}, \bibinfo{person}{Kai Qiu},
  \bibinfo{person}{Jianlong Fu}, {and} \bibinfo{person}{Dongmei Fu}.}
  \bibinfo{year}{2019}\natexlab{}.
\newblock \showarticletitle{Learning recurrent structure-guided attention
  network for multi-person pose estimation}. In
  \bibinfo{booktitle}{\emph{ICME}}. IEEE, \bibinfo{pages}{418--423}.
\newblock


\bibitem[\protect\citeauthoryear{Qiu, Qiu, Fu, and Fu}{Qiu
  et~al\mbox{.}}{2020}]%
        {qiu2020dgcn}
\bibfield{author}{\bibinfo{person}{Zhongwei Qiu}, \bibinfo{person}{Kai Qiu},
  \bibinfo{person}{Jianlong Fu}, {and} \bibinfo{person}{Dongmei Fu}.}
  \bibinfo{year}{2020}\natexlab{}.
\newblock \showarticletitle{Dgcn: Dynamic graph convolutional network for
  efficient multi-person pose estimation}. In \bibinfo{booktitle}{\emph{AAAI}},
  Vol.~\bibinfo{volume}{34}. \bibinfo{pages}{11924--11931}.
\newblock


\bibitem[\protect\citeauthoryear{Rogez, Weinzaepfel, and Schmid}{Rogez
  et~al\mbox{.}}{2017}]%
        {rogez2019lcr}
\bibfield{author}{\bibinfo{person}{Gregory Rogez}, \bibinfo{person}{Philippe
  Weinzaepfel}, {and} \bibinfo{person}{Cordelia Schmid}.}
  \bibinfo{year}{2017}\natexlab{}.
\newblock \showarticletitle{Lcr-net: Localization-classification-regression for
  human pose}.
\newblock  (\bibinfo{year}{2017}), \bibinfo{pages}{3433--3441}.
\newblock


\bibitem[\protect\citeauthoryear{Rogez, Weinzaepfel, and Schmid}{Rogez
  et~al\mbox{.}}{2019}]%
        {rogez2019lcr++}
\bibfield{author}{\bibinfo{person}{Gregory Rogez}, \bibinfo{person}{Philippe
  Weinzaepfel}, {and} \bibinfo{person}{Cordelia Schmid}.}
  \bibinfo{year}{2019}\natexlab{}.
\newblock \showarticletitle{Lcr-net++: Multi-person 2d and 3d pose detection in
  natural images}.
\newblock \bibinfo{journal}{\emph{TPAMI}} \bibinfo{volume}{42},
  \bibinfo{number}{5} (\bibinfo{year}{2019}), \bibinfo{pages}{1146--1161}.
\newblock


\bibitem[\protect\citeauthoryear{Sun, Xiao, Liu, and Wang}{Sun
  et~al\mbox{.}}{2019}]%
        {sun2019deep}
\bibfield{author}{\bibinfo{person}{Ke Sun}, \bibinfo{person}{Bin Xiao},
  \bibinfo{person}{Dong Liu}, {and} \bibinfo{person}{Jingdong Wang}.}
  \bibinfo{year}{2019}\natexlab{}.
\newblock \showarticletitle{Deep high-resolution representation learning for
  human pose estimation}. In \bibinfo{booktitle}{\emph{CVPR}}.
  \bibinfo{pages}{5693--5703}.
\newblock


\bibitem[\protect\citeauthoryear{Sun, Shang, Liang, and Wei}{Sun
  et~al\mbox{.}}{2017}]%
        {sun2017compositional}
\bibfield{author}{\bibinfo{person}{Xiao Sun}, \bibinfo{person}{Jiaxiang Shang},
  \bibinfo{person}{Shuang Liang}, {and} \bibinfo{person}{Yichen Wei}.}
  \bibinfo{year}{2017}\natexlab{}.
\newblock \showarticletitle{Compositional human pose regression}. In
  \bibinfo{booktitle}{\emph{ICCV}}. \bibinfo{pages}{2602--2611}.
\newblock


\bibitem[\protect\citeauthoryear{Sun, Xiao, Wei, Liang, and Wei}{Sun
  et~al\mbox{.}}{2018}]%
        {sun2018integral}
\bibfield{author}{\bibinfo{person}{Xiao Sun}, \bibinfo{person}{Bin Xiao},
  \bibinfo{person}{Fangyin Wei}, \bibinfo{person}{Shuang Liang}, {and}
  \bibinfo{person}{Yichen Wei}.} \bibinfo{year}{2018}\natexlab{}.
\newblock \showarticletitle{Integral human pose regression}. In
  \bibinfo{booktitle}{\emph{ECCV}}. \bibinfo{pages}{529--545}.
\newblock


\bibitem[\protect\citeauthoryear{Tian, Chen, and Shen}{Tian
  et~al\mbox{.}}{2019}]%
        {tian2019directpose}
\bibfield{author}{\bibinfo{person}{Zhi Tian}, \bibinfo{person}{Hao Chen}, {and}
  \bibinfo{person}{Chunhua Shen}.} \bibinfo{year}{2019}\natexlab{}.
\newblock \showarticletitle{Directpose: Direct end-to-end multi-person pose
  estimation}.
\newblock \bibinfo{journal}{\emph{arXiv preprint arXiv:1911.07451}}
  (\bibinfo{year}{2019}).
\newblock


\bibitem[\protect\citeauthoryear{Wang, Li, Liu, Qian, and Lu}{Wang
  et~al\mbox{.}}{2020}]%
        {wang2020hmor}
\bibfield{author}{\bibinfo{person}{Can Wang}, \bibinfo{person}{Jiefeng Li},
  \bibinfo{person}{Wentao Liu}, \bibinfo{person}{Chen Qian}, {and}
  \bibinfo{person}{Cewu Lu}.} \bibinfo{year}{2020}\natexlab{}.
\newblock \showarticletitle{Hmor: Hierarchical multi-person ordinal relations
  for monocular multi-person 3d pose estimation}. In
  \bibinfo{booktitle}{\emph{ECCV}}. Springer, \bibinfo{pages}{242--259}.
\newblock


\bibitem[\protect\citeauthoryear{Wang, Qiu, Peng, Fu, and Zhu}{Wang
  et~al\mbox{.}}{2019}]%
        {wang2019ai}
\bibfield{author}{\bibinfo{person}{Jianbo Wang}, \bibinfo{person}{Kai Qiu},
  \bibinfo{person}{Houwen Peng}, \bibinfo{person}{Jianlong Fu}, {and}
  \bibinfo{person}{Jianke Zhu}.} \bibinfo{year}{2019}\natexlab{}.
\newblock \showarticletitle{AI coach: Deep human pose estimation and analysis
  for personalized athletic training assistance}. In
  \bibinfo{booktitle}{\emph{ACM MM}}. \bibinfo{pages}{374--382}.
\newblock


\bibitem[\protect\citeauthoryear{Wehrbein, Rudolph, Rosenhahn, and
  Wandt}{Wehrbein et~al\mbox{.}}{2021}]%
        {wehrbein2021probabilistic}
\bibfield{author}{\bibinfo{person}{Tom Wehrbein}, \bibinfo{person}{Marco
  Rudolph}, \bibinfo{person}{Bodo Rosenhahn}, {and} \bibinfo{person}{Bastian
  Wandt}.} \bibinfo{year}{2021}\natexlab{}.
\newblock \showarticletitle{Probabilistic monocular 3d human pose estimation
  with normalizing flows}. In \bibinfo{booktitle}{\emph{ICCV}}.
  \bibinfo{pages}{11199--11208}.
\newblock


\bibitem[\protect\citeauthoryear{Wei, Sun, Li, Wang, and Lin}{Wei
  et~al\mbox{.}}{2020}]%
        {wei2020point}
\bibfield{author}{\bibinfo{person}{Fangyun Wei}, \bibinfo{person}{Xiao Sun},
  \bibinfo{person}{Hongyang Li}, \bibinfo{person}{Jingdong Wang}, {and}
  \bibinfo{person}{Stephen Lin}.} \bibinfo{year}{2020}\natexlab{}.
\newblock \showarticletitle{Point-set anchors for object detection, instance
  segmentation and pose estimation}. In \bibinfo{booktitle}{\emph{ECCV}}.
  Springer, \bibinfo{pages}{527--544}.
\newblock


\bibitem[\protect\citeauthoryear{Xiao, Wu, and Wei}{Xiao et~al\mbox{.}}{2018}]%
        {xiao2018simple}
\bibfield{author}{\bibinfo{person}{Bin Xiao}, \bibinfo{person}{Haiping Wu},
  {and} \bibinfo{person}{Yichen Wei}.} \bibinfo{year}{2018}\natexlab{}.
\newblock \showarticletitle{Simple baselines for human pose estimation and
  tracking}. In \bibinfo{booktitle}{\emph{ECCV}}. \bibinfo{pages}{466--481}.
\newblock


\bibitem[\protect\citeauthoryear{Yan, Xiong, and Lin}{Yan
  et~al\mbox{.}}{2018}]%
        {yan2018spatial}
\bibfield{author}{\bibinfo{person}{Sijie Yan}, \bibinfo{person}{Yuanjun Xiong},
  {and} \bibinfo{person}{Dahua Lin}.} \bibinfo{year}{2018}\natexlab{}.
\newblock \showarticletitle{Spatial temporal graph convolutional networks for
  skeleton-based action recognition}. In \bibinfo{booktitle}{\emph{AAAI}}.
\newblock


\bibitem[\protect\citeauthoryear{Yang, Ouyang, Wang, Ren, Li, and Wang}{Yang
  et~al\mbox{.}}{2018}]%
        {yang20183d}
\bibfield{author}{\bibinfo{person}{Wei Yang}, \bibinfo{person}{Wanli Ouyang},
  \bibinfo{person}{Xiaolong Wang}, \bibinfo{person}{Jimmy Ren},
  \bibinfo{person}{Hongsheng Li}, {and} \bibinfo{person}{Xiaogang Wang}.}
  \bibinfo{year}{2018}\natexlab{}.
\newblock \showarticletitle{3d human pose estimation in the wild by adversarial
  learning}. In \bibinfo{booktitle}{\emph{CVPR}}. \bibinfo{pages}{5255--5264}.
\newblock


\bibitem[\protect\citeauthoryear{Zanfir, Marinoiu, and Sminchisescu}{Zanfir
  et~al\mbox{.}}{2018a}]%
        {zanfir2018monocular}
\bibfield{author}{\bibinfo{person}{Andrei Zanfir}, \bibinfo{person}{Elisabeta
  Marinoiu}, {and} \bibinfo{person}{Cristian Sminchisescu}.}
  \bibinfo{year}{2018}\natexlab{a}.
\newblock \showarticletitle{Monocular 3d pose and shape estimation of multiple
  people in natural scenes-the importance of multiple scene constraints}. In
  \bibinfo{booktitle}{\emph{CVPR}}. \bibinfo{pages}{2148--2157}.
\newblock


\bibitem[\protect\citeauthoryear{Zanfir, Marinoiu, Zanfir, Popa, and
  Sminchisescu}{Zanfir et~al\mbox{.}}{2018b}]%
        {zanfir2018deep}
\bibfield{author}{\bibinfo{person}{Andrei Zanfir}, \bibinfo{person}{Elisabeta
  Marinoiu}, \bibinfo{person}{Mihai Zanfir}, \bibinfo{person}{Alin-Ionut Popa},
  {and} \bibinfo{person}{Cristian Sminchisescu}.}
  \bibinfo{year}{2018}\natexlab{b}.
\newblock \showarticletitle{Deep network for the integrated 3d sensing of
  multiple people in natural images}.
\newblock \bibinfo{journal}{\emph{NeurIPS}}  \bibinfo{volume}{31}
  (\bibinfo{year}{2018}), \bibinfo{pages}{8410--8419}.
\newblock


\bibitem[\protect\citeauthoryear{Zeng, Sun, Huang, Liu, Xu, and Lin}{Zeng
  et~al\mbox{.}}{2020}]%
        {zeng2020srnet}
\bibfield{author}{\bibinfo{person}{Ailing Zeng}, \bibinfo{person}{Xiao Sun},
  \bibinfo{person}{Fuyang Huang}, \bibinfo{person}{Minhao Liu},
  \bibinfo{person}{Qiang Xu}, {and} \bibinfo{person}{Stephen Lin}.}
  \bibinfo{year}{2020}\natexlab{}.
\newblock \showarticletitle{Srnet: Improving generalization in 3d human pose
  estimation with a split-and-recombine approach}. In
  \bibinfo{booktitle}{\emph{ECCV}}. Springer, \bibinfo{pages}{507--523}.
\newblock


\bibitem[\protect\citeauthoryear{Zhao, Peng, Tian, Kapadia, and Metaxas}{Zhao
  et~al\mbox{.}}{2019}]%
        {zhao2019semantic}
\bibfield{author}{\bibinfo{person}{Long Zhao}, \bibinfo{person}{Xi Peng},
  \bibinfo{person}{Yu Tian}, \bibinfo{person}{Mubbasir Kapadia}, {and}
  \bibinfo{person}{Dimitris~N Metaxas}.} \bibinfo{year}{2019}\natexlab{}.
\newblock \showarticletitle{Semantic graph convolutional networks for 3d human
  pose regression}. In \bibinfo{booktitle}{\emph{CVPR}}.
  \bibinfo{pages}{3425--3435}.
\newblock


\bibitem[\protect\citeauthoryear{Zhen, Fang, Sun, Liu, Jiang, Bao, and
  Zhou}{Zhen et~al\mbox{.}}{2020}]%
        {zhen2020smap}
\bibfield{author}{\bibinfo{person}{Jianan Zhen}, \bibinfo{person}{Qi Fang},
  \bibinfo{person}{Jiaming Sun}, \bibinfo{person}{Wentao Liu},
  \bibinfo{person}{Wei Jiang}, \bibinfo{person}{Hujun Bao}, {and}
  \bibinfo{person}{Xiaowei Zhou}.} \bibinfo{year}{2020}\natexlab{}.
\newblock \showarticletitle{Smap: Single-shot multi-person absolute 3d pose
  estimation}. In \bibinfo{booktitle}{\emph{ECCV}}. Springer,
  \bibinfo{pages}{550--566}.
\newblock


\bibitem[\protect\citeauthoryear{Zhou, Wang, and Kr{\"a}henb{\"u}hl}{Zhou
  et~al\mbox{.}}{2019}]%
        {zhou2019objects}
\bibfield{author}{\bibinfo{person}{Xingyi Zhou}, \bibinfo{person}{Dequan Wang},
  {and} \bibinfo{person}{Philipp Kr{\"a}henb{\"u}hl}.}
  \bibinfo{year}{2019}\natexlab{}.
\newblock \showarticletitle{Objects as points}.
\newblock \bibinfo{journal}{\emph{arXiv preprint arXiv:1904.07850}}
  (\bibinfo{year}{2019}).
\newblock


\end{thebibliography}

\newpage
\begin{appendix}
\section*{Supplementary Material}

In this supplementary material, we introduce the algorithm details in Section \ref{sec:alg}. To better understand the dynamic graph reasoning (DGR) in our approach, we show the visualization of the dynamic decoding graph in Section \ref{sec:graph}. The limitations and failure cases of our approaches are shown in Section \ref{sec:failcases}. More visualization cases are shown in Section \ref{sec:cases}.

\section{Algorithm Details}
\label{sec:alg}

The algorithm details of our GR-M3D are shown in Algorithm \ref{alg}. In Algorithm \ref{alg}, the backbone network can be ResNet~\cite{he2016deep}, HRNet~\cite{sun2019deep}, and Hourglass~\cite{newell2016stacked}, etc. Unless otherwise specified, the backbone of GR-M3D is the Hourglass network. $conv$ means the convolutional layers with a kernel size of $1\times 1$ to generate the four data maps.

\begin{algorithm}[h]\small
  \caption{GR-M3D with dynamic graph reasoning (DGR)} 
  \label{alg}
  \begin{algorithmic}[1]
    \Require
      $I$: Input image;
      $K$: Joint number;
      $N$: Person number;
      $\phi(\cdot)$: Backbone network;
      $F$: Deep features;
      $conv$: Convolution layers with kernel size of $1\times 1$;
      $SDAR(\cdot)$: The Scale and Depth Aware Refinement module.

    \Ensure
      $\mathbb{P}_{3d}$: $\{\hat{p}^j_{3d}, j \in [1,K]\}$;\\
      
      $F = \phi(I)$;
      \State $M^I_h, M^I_s, M^I_d, M^I_o = conv(F)$;
      
      \State $M_h, M_s, M_d, M_o = SDAR(M^I_h, M^I_s, M^I_d, M^I_o, F)$;
      
      \State Obtain center points set $C=\{c_1, c_2, ..., c_N\}$ from $M_h$;
      
      \State Obtain 2D keypoints set $P_{2D} = \{p_1, p_2, ..., p_M\}$ from $M_h$;
      
      \State Assign $P_{2D}$ to $C$ as Eq.(6);
      
      \State Obtain dense decoding paths $\mathbb{E}=\{e^{ij}, i\in [1,K], j\in [1,K]\}$;
      
      \State Calculate dynamic decoding graphs $\hat{\mathbb{E}}=\{(e^{ij}, \mathcal{W}(p^i, p^j)), i\in [1,K], j\in [1,K]\}$ according to Eq.(8), Eq.(9), and Eq.(10);
      
      \State Calculate 3D coordinates $\hat{p}^j_{3d}$ as Eq.(11).
      
  \end{algorithmic}
\end{algorithm}

\section{Dynamic Decoding Graph}
\label{sec:graph}

\begin{figure}[t]
\centering
\includegraphics[width=0.95\columnwidth]{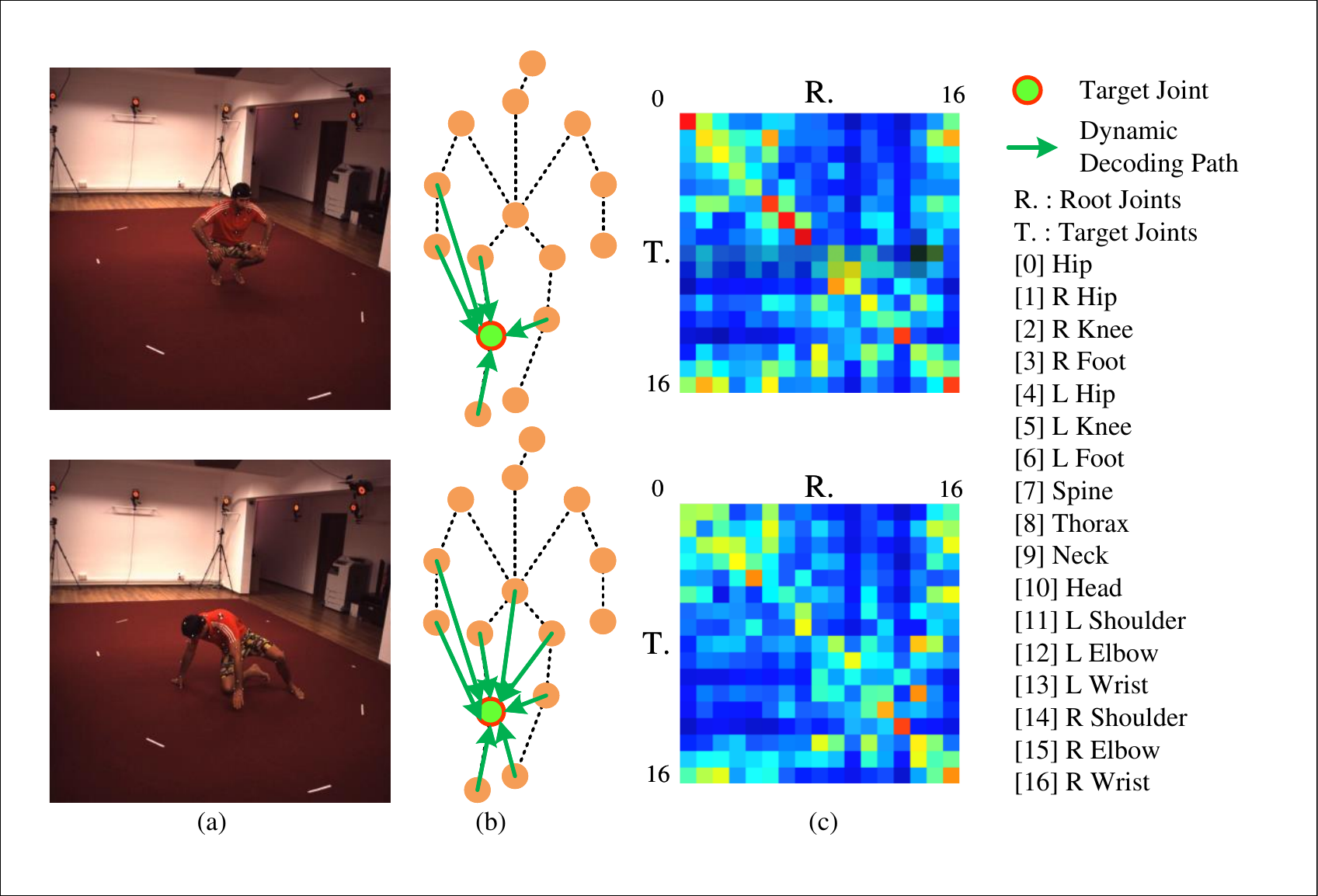} 
\caption{The visualization of dynamic decoding graphs. (a) Different actions from the same person, (b) The illustration of dynamic decoding paths (Green arrows show the decoding paths that weight is greater than $0.1$), (c) The weight matrix of the predicted dynamic decoding graph(DDG). The DDGs for the two cases from the same person are different, which demonstrates that our method can self-adaptively estimate the best decoding graph for each person according to the different inputs.}
\label{fig:graph}
\end{figure}

To better understand the proposed dynamic graph reasoning (DGR) in GR-M3D for 3D human pose estimation, we visualize the path weights in dynamic decoding graphs of different people in Figure \ref{fig:graph}.

As shown in Figure \ref{fig:graph}, for the two input cases performed by the same person, our GR-M3D predicted different decoding graphs as Figure \ref{fig:graph} (c). As shown in Figure \ref{fig:graph} (b), for the cases in the second row, the decoding graph contains more decoding paths with high weights ($>0.1$) from the hip, left hip, and left foot since occlusion. Compared with these two cases, it shows that the predicted decoding graph is self-adapting due to the dynamic graph reasoning (DGR) mechanism.

\begin{figure}[t]
\centering
\includegraphics[width=0.95\columnwidth]{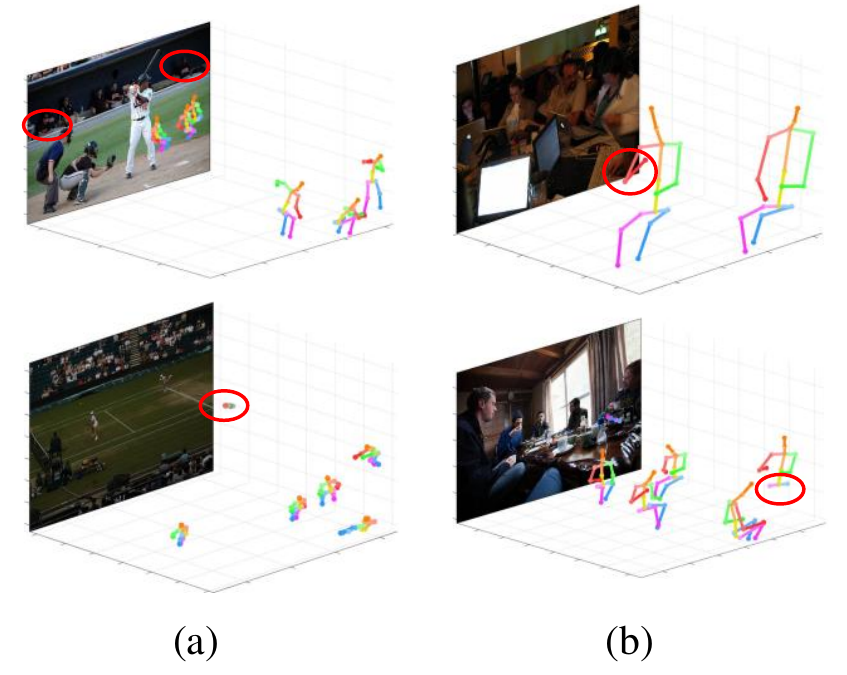} 
\caption{The visualization of failure cases from COCO~\cite{lin2014microsoft} dataset. (a) Some missed instances since they are too small, (b) Some wrong predictions since heavy occlusions.}
\label{fig:failcases}
\end{figure}

\begin{figure*}[]
\centering
\includegraphics[width=\textwidth]{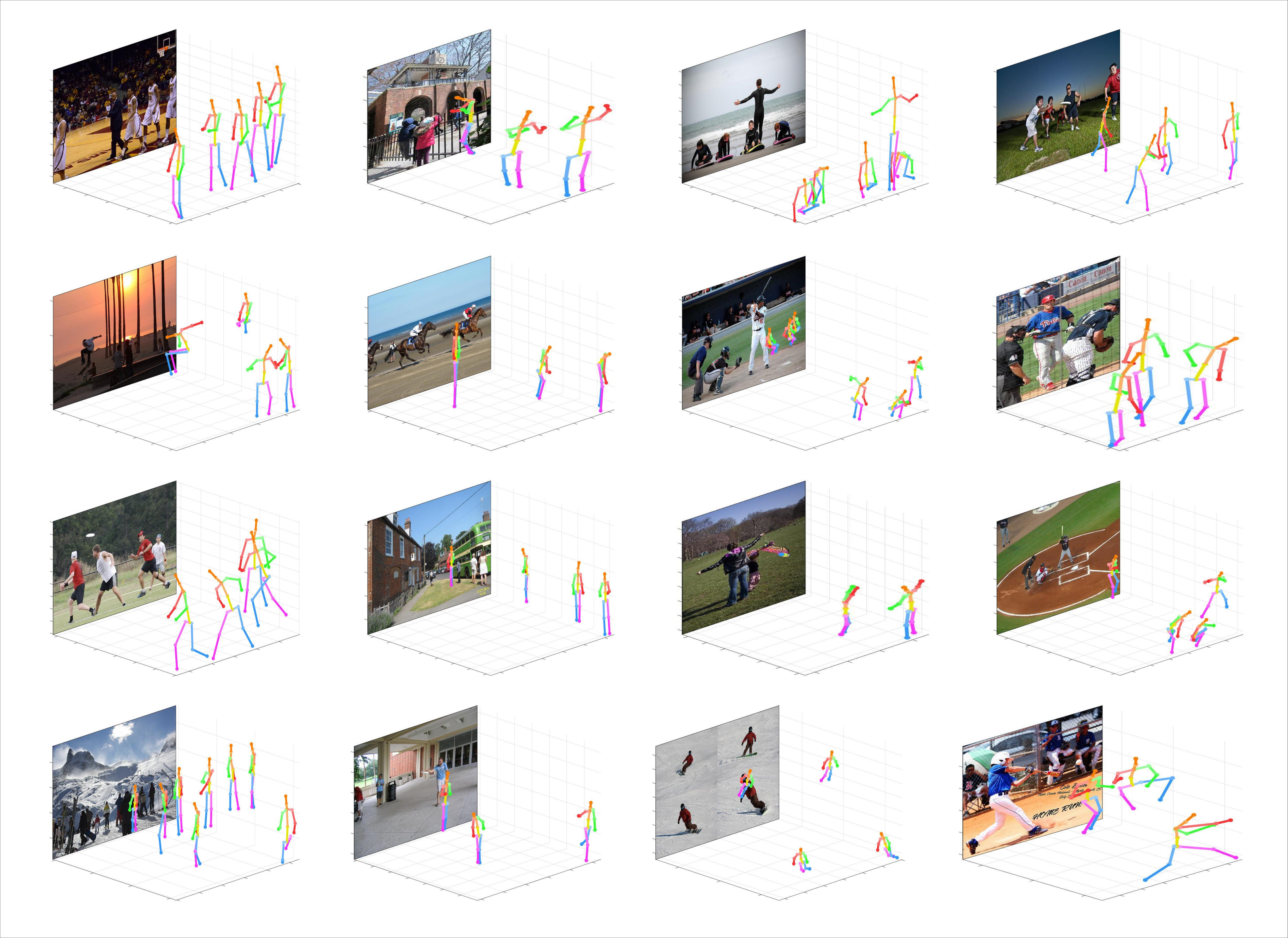} 
\caption{Visualization of the predicted 3D poses by GR-M3D on COCO~\cite{lin2014microsoft} dataset.
}
\label{coco_vis}
\end{figure*}

\section{Limitations and Failure Cases}
\label{sec:failcases}

We discuss the limitations of the proposed GR-M3D in this section and show some failure cases in Figure \ref{fig:failcases}. As shown in Figure \ref{fig:failcases} (a), for some small instances, our approach can not handle and misses them. Due to the input is whole image for GR-M3D, the small instances with limited resolution lack sufficient information for GR-M3D to handle. As shown in Figure \ref{fig:failcases} (b), for some cases with heavy occlusions, it is hard for GM-M3D to tackle.

\section{More Visualization Results}
\label{sec:cases}
\begin{figure*}[t]
\centering
\includegraphics[width=0.95\textwidth]{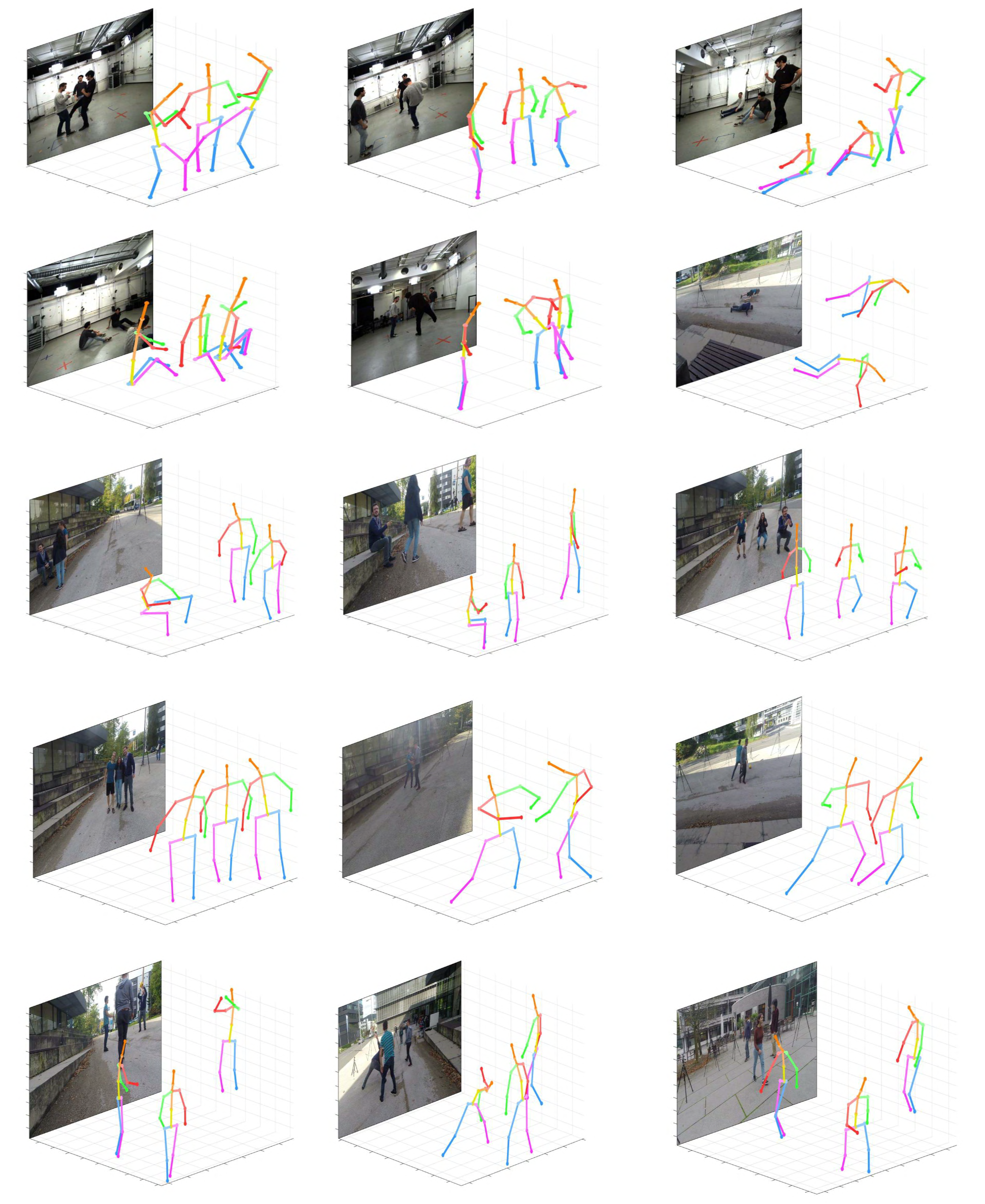} 
\caption{Visualization of the predicted 3D poses by GR-M3D on MuPoTS-3D~\cite{mehta2017monocular} dataset.
}
\label{mu_vis}
\end{figure*}

To demonstrate the generalization ability of proposed GR-M3D, the more visualization results on COCO~\cite{lin2014microsoft} and MuPoTS-3D~\cite{mehta2017monocular} datasets are shown in Figure \ref{coco_vis} and Figure \ref{mu_vis}. The images in COCO~\cite{lin2014microsoft} dataset are collected in unconstraint and in-the-wild conditions. There is no 3D pose annotations in COCO dataset.
The images in MuPoTS-3D~\cite{mehta2017monocular} dataset are collected in constraint and in-the-wild conditions. All the results in Figure \ref{coco_vis} and Figure \ref{mu_vis} are predicted by GR-M3D, which is trained on MuCo-3DHP dataset~\cite{mehta2017monocular}.

As shown in Figure \ref{coco_vis}, our GR-M3D performs well on these in-the-wild images with strange poses, occlusions, and variational backgrounds. As shown in Figure \ref{mu_vis}, our GR-M3D still can handle these images with different human actions, changing camera viewpoint, and occlusions. These results show that our GR-M3D has a strong generalization ability on handling these in-the-wild cases.

\end{appendix}

\end{document}